\RequirePackage{fix-cm}
\documentclass[a4paper,english,aps,prd,twocolumn,10pt,superscriptaddress,nofootinbib,nobibnotes,altaffillsymbol,showpacs,showkeys,floatfix]{revtex4-2}

\usepackage[T1]{fontenc}
\usepackage[utf8]{inputenc}
\usepackage{babel}

\usepackage{amsmath}
\usepackage{amssymb}
\usepackage{mathrsfs}
\usepackage{esint} 
\usepackage{bbm} 

\usepackage[caption=false]{subfig}
\usepackage[pdftex]{graphicx}

\graphicspath{{}}

\usepackage{tabularx}

\usepackage[unicode=true,
 bookmarks=true,bookmarksnumbered=false,bookmarksopen=true,bookmarksopenlevel=2,
 breaklinks=true,pdfborder={0 0 0},backref=false,colorlinks=true]
 {hyperref}

\hypersetup{
	pdftitle={The Discrete Langevin Machine: Bridging the Gap Between Thermodynamic and Neuromorphic Systems},
	pdfauthor={Kades,Pawlowski},
	pdfkeywords={langevin dynamics} {discrete systems} {boltzmann machine} {neuromorphic systems} {lif neurons} {brainscales},
	linkcolor=BlueViolet,anchorcolor=BlueViolet,citecolor=BlueViolet,filecolor=BlueViolet,urlcolor=BlueViolet
}

\allowdisplaybreaks
\usepackage{color}
\usepackage[usenames,dvipsnames, table]{xcolor}

\def\eq#1{(\ref{#1})}

\def\s0#1#2{\mbox{\small{$ \frac{#1}{#2} $}}}
\def\0#1#2{\frac{#1}{#2}}

\definecolor{darkblue}{rgb}{0.0,0.0,0.4}
\definecolor{deepred}{rgb}{0.6,0,0}
\definecolor{deepgreen}{rgb}{0,0.3,0}

\usepackage{array,booktabs}
\newcolumntype{P}[1]{>{\centering\arraybackslash}m{#1}}

\usepackage[subpreambles=true]{standalone}
\usepackage{layouts}

\begin{document}

\title{The Discrete Langevin Machine:\\ Bridging the Gap Between Thermodynamic
  and Neuromorphic Systems}

\author{Lukas Kades}

\affiliation{Institut für Theoretische Physik, Universität Heidelberg,
  Philosophenweg 16, 69120 Heidelberg, Germany}

\author{Jan M. Pawlowski}

\affiliation{Institut für Theoretische Physik, Universität Heidelberg,
  Philosophenweg 16, 69120 Heidelberg, Germany}


\begin{abstract}
  A formulation of Langevin dynamics for discrete systems is derived
  as a class of generic stochastic processes. The dynamics
  simplify for a two-state system and suggest a network
  architecture which is implemented by the Langevin machine. The
  Langevin machine represents a promising approach to compute
  successfully quantitative exact results of Boltzmann distributed
  systems by LIF neurons. Besides a detailed introduction of the
  dynamics, different simplified models of a neuromorphic hardware
  system are studied with respect to a control of emerging sources of
  errors.
\end{abstract}
\maketitle


\section{Introduction}
\label{sec:intro}

The rapidly increasing progress on neuromorphic computing and the
ongoing research of spiking systems as the third generation of
neural networks calls for a better understanding of fundamental
processes of neuromorphic hardware systems~\cite{Maas, Huang, Rezende,
  Vladimir, Marblestone, Gerstner2012}; for a recent review on
neuromorphic computing see~\cite{Schuman}. As a parallel computing
platform, these systems may be used in the long run to accurately
simulate and compute large systems, in particular given their low
energy consumption. 

Possible applications range from an effective implementation of
artificial neural networks and further machine learning
methods~\cite{Maass2014,Emre,Emre2016, Pedroni,Leng,Hinton:2002} over a
better understanding of biological processes in our
brains~\cite{Pecevski, Nessler2013} to the computation of physical and
stochastic interesting systems~\cite{Czischek2018, Carleo2017,
  Gao2017}. Many artificial neural networks and physical systems are
described by Boltzmann distributed systems. For a quantitatively
accurate computation of such systems, it is necessary to deduce an
exact representation on neuromorphic hardware
systems~\cite{Jordan2015, Pedroni, Probst, Buesing2011}, in particular
for systematic error estimates.

Our work is motivated by the similarity of Langevin dynamics and
leaky integrate-and-fire (LIF) neurons for performing stochastic
inference~\cite{Petrovici2016}. Indeed, the fundamental dynamics of
LIF neurons is governed by Langevin dynamics. Apart from its obvious
relevance for the description of stochastical processes, the Langevin
equation~\cite{Langevin1908} can also be used for simulating quantum
field theories with stochastic
quantization~\cite{Damgaard:1987rr,Batrouni,Parisi1981}.  In this
approach the Euclidean path integral measure is obtained as the
stationary distribution of a stochastic process. This paves the way to
the heuristic approach of using complex Langevin dynamics as a
potential method for accessing real time dynamics and sign
  problems. The latter problem is, e.g.\ prominent in QCD at finite
chemical potential~\cite{Aarts2008, Aarts2012, Aarts2013}. A further
interesting application of the Langevin equation can be found
in~\cite{Welling2011}. Langevin dynamics is combined with a
stochastic gradient descent algorithm to perform Bayesian learning
which enables an uncertainty estimation of resulting parameters.

Many of the above mentioned systems are discrete ones, the simplest
one being a two-state system. This suggests the formulation of a
discrete analog to continuous Langevin dynamics, for the accurate
description of discrete systems. In the present work, we show that a
formulation of Langevin dynamics for discrete systems leads to a class of a generic stochastic process, namely, the \textit{Langevin
  equation for discrete systems},
\begin{equation}
\label{eq:updaterulesimplee}
\phi'=\phi+(\nu-\phi)\Theta\left[-1-\frac{\epsilon}{2\lambda_\epsilon}
  \Delta S(\nu,\phi)+\sqrt{\epsilon}\tilde{\eta}\right]\,,
\end{equation}
where $\phi$ is the current state and
$\phi'$ the updated state. The process is driven by a Gaussian noise term $\tilde{\eta}$. The parameter $\epsilon$ has an impact on the acceptance probability of the proposed state $\nu$ and should be chosen small. The term $\Delta S(\nu, \phi):=S(\nu) - S(\phi)$ measures the change of the action $S$ of the system under a transition from state $\phi$ to the proposed state $\nu$. A more detailed derivation of
\eq{eq:updaterulesimplee} including a discussion of its properties can
be found in Section~\ref{sec:langevindynamicsdiscrete}. The dynamics leads in the limit of $\epsilon\to0$ to Boltzmann distributed states.

\begin{figure*}
	\centering
	\includegraphics[width=\linewidth]{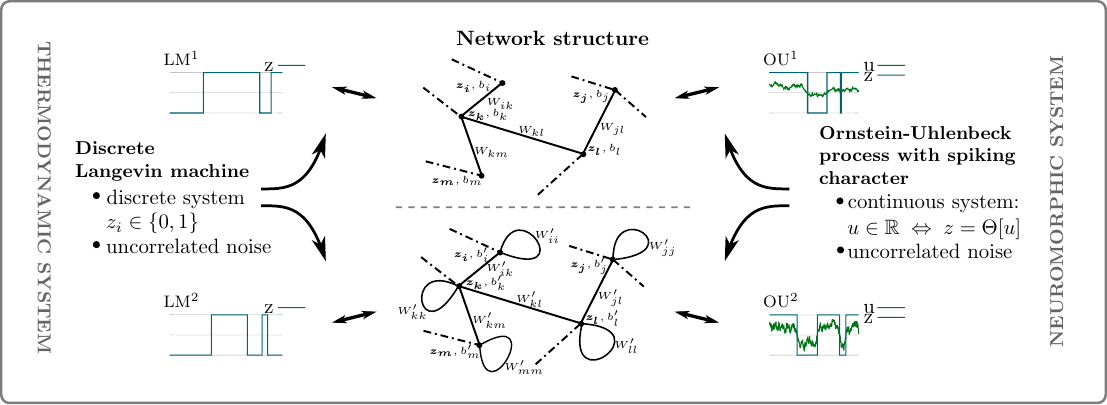}
	\caption{Comparison of the commonly used network structure
          (upper row) and the presented architecture with a
          self-interacting contribution (lower row). Both network
          structures can be considered as systems of two discrete
          states with an uncorrelated noise contribution, which
          corresponds to different implementations of the discrete
          Langevin machine. Their continuous counterpart is
          represented by an Ornstein-Uhlenbeck process with spiking
          character. The dynamics is based on the temporal evolution of
          a membrane potential $u:=u_\textnormal{eff}(t)$. The
          interaction of neurons relies on a projection of the
          potential onto two states and enables a comparison with the
          Langevin machine. The processes on the right-hand side are
          already very close to the fundamental dynamics of LIF
          sampling.}
	\label{fig:systems}
\end{figure*}

The present work concentrates on the potential of the process
for a more accurate implementation of Boltzmann distributed systems on
the neuromorphic hardware. This leads to an architecture of neurons
based on a self-interacting contribution. The self-interacting
  term changes manifestly the dynamics of the neural network. This
  results in an activation function which is much closer to a logistic
  distribution, the activation function of a Boltzmann machine, than
  existing approaches. The architecture can be applied to both,
  discrete two-state systems and neuromorphic hardware systems with a
  continuous membrane potential and a spiking character. The dynamics differ in their kind of noise that is
  uncorrelated in the former case and autocorrelated in the latter
  case. In this work
  \textit{spiking character} refers to an effective mapping of a
  continuous potential to two discrete neuron states in an interacting
  system. Figure~\ref{fig:systems} compares the different network
  structures and gives an overview over existing and contributed dynamics of this work.

An exact representation of the activation function of the Boltzmann
machine is necessary to obtain correct statistics in coupled systems
on neuromorphic hardware systems. In the present work we show in a
detailed numerical analysis that small deviations in the activation
function propagate if a rectangular refractory mechanism or
interactions between neurons are taken into account. These small
deviations have a large impact on the resulting correlation functions and
observables. The numerical results demonstrate that a reliable
estimation, an understanding and a control of different sources of errors
are essential for a correct computation of Boltzmann distributed
systems in the future.

An introduction to Langevin dynamics is given in Section~\ref{sec:metrvslang}. The Langevin equation for discrete
systems is derived in Section~\ref{sec:langevindynamicsdiscrete}. In
Section~\ref{sec:langevinmachine}, the so-called
\textit{sign-dependent discrete Langevin machine} is introduced as a
special case of the Langevin equation for discrete systems. The
mapping of different dynamics of discrete systems onto an
Ornstein-Uhlenbeck process with spiking character is discussed in
Section~\ref{sec:neuromorphichardware}. Section~\ref{sec:wholepicture}
recapitulates relations between the discrete Langevin machine and the
neuromorphic hardware. In Section~\ref{sec:applications},
numerical results of the introduced and of existing dynamics are
presented, possible sources of errors are extracted and the
propagation of errors for different abstractions of a neuromorphic
hardware system is analyzed. The conclusion and outlook can be found
in Section~\ref{sec:concl}.

\section{Metropolis Algorithm Versus Langevin Dynamics}
\label{sec:metrvslang}

The section reviews shortly properties of the Metropolis
algorithm~\cite{Metropolis} and Langevin dynamics for continuous
systems to get a first intuition on how a possible Langevin equation
for a discrete system could look like. The considerations are inspired
by the work in~\cite{Baillie, Meakin, Ettelaie}.

\subsection{Langevin dynamics}

We adopt the common formulation of the Langevin equation within the
study of Euclidean quantum field
theories~\cite{Damgaard:1987rr,Batrouni},
\begin{equation}
\label{eq:langevin}
\frac{\partial}{\partial \tau} \phi_x(\tau)=-\frac{\delta S_E}{\delta \phi_x(\tau)}+\eta_x(\tau)\,,
\end{equation}
where $S_E$ corresponds to the Euclidean action, which depends on
fields $\phi_x(\tau)$ on a (3+1)-dimensional hypercubic lattice in
Euclidean space. The Langevin equation describes the evolution of the
quantum fields $\phi_x(\tau)$ in an additional fictious time
dimension, the Langevin time $\tau$. Quantum fluctuations are emulated
by the additional white Gaussian noise term with the properties to be
uncorrelated,
\begin{equation}
\langle \eta_i, \eta'_j \rangle_{\eta} = 2\delta(j-i)\delta(t'-t)\,,\quad\langle\eta_i\rangle_{\eta}=0\,.
\end{equation}
It can be shown that in equilibrium the resulting distribution of the
fields coincides with the Boltzmann distribution:
$\lim\limits_{\tau\to\infty} P(\phi, t) = P_\textnormal{eq}(\phi) =
\frac1Z \exp(-S_E)$. A common approach to prove this is to derive the
equivalence of the Langevin equation and the Fokker-Planck equation in
a first step and to compute the static solution in a second
step~\cite{Damgaard:1987rr}. This property renders Langevin dynamics a
powerful tool in QCD~\cite{Aarts2008, Aarts2012} and beyond.
\subsection{Equivalence to a Monte Carlo algorithm}
\label{sec:equivalence}

The transition probability $W(\phi\to\phi')$ for Langevin dynamics is derived in Appendix~\ref{sec:transitionprobability}. It is computed based on a discrete form of the Langevin equation where an infinitesimal update step of a field $\phi:=\phi(\tau)$ to a field $\phi':=\phi(\tau+\epsilon)$ is considered. The resulting transition probability can be rewritten as
\begin{equation}
W(\phi\to\phi')=\frac{1}{\sqrt{2\epsilon}}\varphi\left(\frac{\phi'-
	\phi}{\sqrt{2\epsilon}}\right)\exp\left[-\frac{S(\phi') - S(\phi)}{2}\right]\,.
\end{equation}
The first factor can be interpreted as the selection probability of a field and the second term as the acceptance probability. There are
two adaptations to the Metropolis algorithm~\cite{Metropolis}. First, the acceptance
probability has the additional factor of $2$. This factor is necessary
to satisfy the detailed balance equation. Second, the proposal state
is chosen to be an infinitesimal change to the current state. This
ensures a correct normalization of the transition probability for
$\epsilon\to 0$, as can be seen by replacing the selection probability
by the delta distribution,
\begin{align}
\int_{-\infty}^{\infty}& d\phi' W(\phi\to\phi') =\nonumber\\[1ex] &=
\int_{-\infty}^{\infty} d\phi'
\delta(\phi'-\phi)\exp\left[-\frac{S(\phi') - S(\phi)}{2}\right]=1\,.
\end{align}
The update mechanism of Langevin dynamics can be parallelized
because of these two properties.

Accordingly, the Langevin equation~(\ref{eq:langevin}) can be
interpreted as a standard Monte Carlo algorithm with a Gaussian
distribution as selection probability. The proposal state is chosen
implicitly by an absorption of the acceptance probability into the
selection probability and by a corresponding sampling with Gaussian
noise. Since the nearest neighbor sites can be assumed to be nearly
constant in one Monte Carlo step, it is possible to switch from a
random sequential update formalism to a parallel update of the entire
lattice. The Langevin time is introduced as a temporal measure for a
lattice update. In principle, the delta distribution can be exchanged
by any other positive representation. In some cases one can derive
Langevin dynamics with another noise source by an equivalent
approximation (for example, Cauchy noise).

Having the knowledge of this section, a Langevin equation for discrete
systems can be constructed by considering a standard Monte Carlo
algorithm.

\section{Discrete Langevin Dynamics}
\label{sec:langevindynamicsdiscrete}
The formulation of the \textit{Langevin equation for discrete systems} is derived
inspired by the comparison of the Metropolis algorithm and Langevin
dynamics in the previous section. The general formulation of a Langevin
equation for discrete systems presented in this section is
similar to a Monte Carlo algorithm and is driven
by a Gaussian noise contribution. The transition probability to a
proposed state is regulated by the introduction of truncating
Gaussian noise. It is shown that the accuracy of the process strongly
depends on an intrinsic parameter $\epsilon$ and the scale of the
energy contribution.

Certain necessary properties of a possible Langevin equation for
discrete systems can be stated beforehand based on the comparison of Langevin dynamics and the Metropolis algorithm in
Section~\ref{sec:metrvslang}. First, an infinitesimal change
of the microscopic state/field is not possible. Therefore, one has to
switch from a parallel to a random sequential update
mechanism. Further, the computer time is the timescale of
the approach. The proposal field has to be chosen from a discrete
distribution. One may select the proposal field according to some
distribution around the current field. However, since a
parallelization is not possible, the uniform selection probability of
a Metropolis algorithm can be adopted.

Assuming the same acceptance probability as in the continuous case, a
starting point is the following proportionality:
\begin{equation}
\label{eq:transitionprobability}
W(\phi\to\phi')\propto\exp\left[-\frac{S(\phi') - S(\phi)}{2}\right]\,.
\end{equation}
With the help of a relation between the cumulative Gaussian
distribution $\Phi(x)$ and the exponential function, given by~(\ref{eq:cumexp}),
and with $\Delta S(\phi',\phi)=S(\phi')-S(\phi)$, this can be
rewritten in the following way:
\begin{align}
  W(\phi\to\phi')&\propto \Phi\left(-\frac{1}{\sqrt{\epsilon}}-
                   \frac{\sqrt{\epsilon}}{2\lambda_\epsilon}
                   \Delta S(\phi',\phi)\right)=\nonumber\\[1ex]&
                                                                 =P\left(\tilde{\eta} <
                                                                 -\frac{1}{\sqrt{\epsilon}}
                                                                 -\frac{\sqrt{\epsilon}}{2\lambda_\epsilon}\Delta S(\phi',\phi)\right)\,,
\end{align}
for $\epsilon\to 0$. An analytical expression for the additional
scaling factor $\lambda_\epsilon$ is given in
equation~(\ref{eq:lambda}).

The Gaussian noise contribution $\tilde{\eta}$ is uncorrelated and has variance $1$, 
\begin{equation}
  \langle \tilde{\eta}_i, \tilde{\eta}'_j \rangle_{\tilde{\eta}} = \delta(j-i)\delta(t'-t)\,,
  \qquad\langle\tilde{\eta}_i\rangle_{\tilde{\eta}}=0\,.
\end{equation}
Taking the current state $\phi$ into account, one can transform the
sampling from the cumulative normal distribution into a general
stochastic update rule with Gaussian noise and a proposal state
$\nu$. This leads us to \eq{eq:updaterulesimplee}, which was already presented in
the introduction, 
\begin{equation}
\label{eq:updaterulesimple}
\phi'=\phi+(\nu-\phi)\Theta\left[-1-\frac{\epsilon}{2
    \lambda_\epsilon}\Delta S(\nu,\phi)+\sqrt{\epsilon}\tilde{\eta}\right]\,,
\end{equation}
where $\epsilon$ needs to be chosen sufficiently small. $\Theta(x)$
represents the Heaviside function. 

The update formalism corresponds to a single spin flip Monte Carlo
algorithm with a random sequential update mechanism, driven by
Gaussian noise. It can be immediately
seen within the present form that a flip to a proposed field gets the
more unlikely the smaller $\epsilon$. Adaptations of the Gaussian
noise term to truncated Gaussian noise can help to improve the
dynamics, i.e., to increase the probability of a spin flip. In
principle, this corresponds to a rescaling of the transition
probability term similar to a maximization of the spin flip
probability in the Metropolis algorithm.

The truncated Gaussian noise term can be expressed by the following
parametrization:
\begin{equation}
\label{eq:truncatednoise}
\tilde{\eta}^T \in \left[\frac{1}{\sqrt{\epsilon}} + \alpha, \infty \right]\,,
\end{equation}
where $\alpha$ is in the range of 
\begin{equation}
  -\infty \leq \alpha \leq -\frac{\sqrt{\epsilon}}{2\lambda_\epsilon}
  \Delta_{\max}\,,\quad\textnormal{with}\quad\Delta_{\max}=|\Delta S(\nu,\phi)|\,.
\end{equation}
The improved update rule is
\begin{equation}
\label{eq:updateformalismtruncated}
\phi'=\phi+(\nu-\phi)\Theta\left[-1-
  \frac{\epsilon}{2\lambda_\epsilon}\Delta S(\nu,\phi)
  +\sqrt{\epsilon}\tilde{\eta}^T\right]\,.
\end{equation}
For $\alpha\to-\infty$ this reduces to the update
formalism~(\ref{eq:updaterulesimple}) and for
$\alpha=-\frac{\sqrt{\epsilon}}{2\lambda_\epsilon} \Delta_{\max}$ one
obtains spin flip probabilities up to $1$. This can be seen under
consideration of the explicit transition probability of the update
rule~(\ref{eq:updateformalismtruncated}),
\begin{equation}
  W(\phi\to\nu)=\frac{\Phi\left(-\frac{1}{\sqrt{\epsilon}}-
      \frac{\sqrt{\epsilon}}{2\lambda_\epsilon}\Delta S(\nu,
      \phi)\right)}{\Phi\left(-\frac{1}{\sqrt{\epsilon}}-\alpha\right)}\,.
\end{equation}
Transition probabilities of further standard Monte Carlo algorithms
can be emulated by other choices of $\alpha$. Note that for a uniform
random number $r\in[0,1[$ and a proposal field $\nu$, an equivalent
formulation to~(\ref{eq:updateformalismtruncated}) can be stated for
the transition
probability~(\ref{eq:transitionprobability}), 
\begin{equation}
  \phi'=\phi+(\nu-\phi)\Theta\left[\exp\left(-\frac{S(\nu)-S(\phi)
        +\Delta_{\max}}2\right)-r\right]\,.
\end{equation}
Processes with a different value of $\alpha$, i.e., a different
rescaling of the transition probability, can always be mapped onto
each other by a respective rescaling of the time. Given a transition
probability $W(\phi\to\mu)$ and a scaling factor $a$, the following
relation holds:
\begin{equation}
\label{eq:rescalingrelation}
W(\phi\to\mu)\to a W(\phi\to\mu) \quad
\Leftrightarrow \quad t \to \frac{t}{a}\,.
\end{equation}
Most of the existing single spin flip algorithms can be reformulated
into a Langevin equation for discrete systems
with the same derivation, as presented in this section. However, it
can be shown that for the particular choice of the transition
probability according to equation~(\ref{eq:transitionprobability}),
the resulting order of accuracy in the detailed balance equation is
the best one, with
$\mathcal{O}\left(\epsilon \Delta S(\nu, \phi)^3\right)$. In general, it holds for the presented dynamics that
\begin{equation}
\lim\limits_{\epsilon\to0}P_\textnormal{eq}(\phi)\propto \exp\left[-S(\phi)\right]\,.
\end{equation}

The update formalism~(\ref{eq:updateformalismtruncated}) represents a
Langevin-like equivalent for discrete systems to Langevin dynamics
of continuous systems. As for continuous systems, the dynamics depends
on Gaussian noise and is based on a rather simple
expression. The algorithms can also be applied to continuous systems
due to the equivalence to standard Monte Carlo algorithms in the limit
$\epsilon\to 0$.

\section{Sign-Dependent Discrete Langevin Machine}
\label{sec:langevinmachine}
The Langevin equation for discrete
systems~(\ref{eq:updateformalismtruncated}) turns into a rather simple
expression for a two-state system. The resulting dynamics is
introduced in the following as \textit{sign-dependent discrete
  Langevin machine} ($\textnormal{LM}^2$). The $\textnormal{LM}^2$
represents a architecture for interacting neurons with the
particularity of a self-interacting contribution. The derived
network structure results in a basic dynamics with
different weights and biases compared to the Boltzmann machine. It has
the unique property that the equilibrium distribution converges in the
limit $\epsilon\to 0$, despite a different underlying dynamics, to a
logistic distribution, the activation function of the Boltzmann
machine.

We define the energy of the Boltzmann machine in the
common way by 
\begin{equation}
E = -\sum_{i<j} W_{ij} z_i z_j - \sum_i b_i z_i\,,
\end{equation}
where $W_{ij}$ are symmetric weights between the neurons $i$ and $j$
and $b_i$ is some additional bias. The domain of definition of the
states $z_i$ at each neuron is given by $z_i\in\lbrace 0, 1\rbrace$.
\begin{figure}
	{\resizebox{0.48\linewidth}{!}{\includegraphics[width=0.48\linewidth]{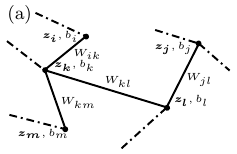}}}\hfill
	{\resizebox{0.48\linewidth}{!}{\includegraphics[width=0.48\linewidth]{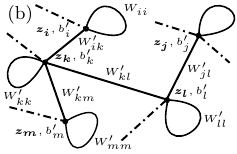}}}
	\caption{Comparison of the structure of a Boltzmann machine~(a) and that of the sign-dependent discrete Langevin machine~(b). The $\textnormal{LM}^2$ has a self-interacting term and rescaled weights and biases. Nevertheless the dynamics leads in equilibrium to a Boltzmann distribution.}
	\label{fig:machines}
\end{figure}

For applying the generalized update
rule~(\ref{eq:updateformalismtruncated}) we need the following
identifications: $S \to E$ and $\phi_i \to z_i$. As discussed in
Appendix~\ref{sec:derivationlangevinmachine}, the following simplified
update rule can be derived for the $\textnormal{LM}^2$:
\begin{equation}
\label{eq:langevinmachine}
z'_i=\Theta\left[W'_{ii}z_i+\sum_{j} W'_{ij}z_j + b'_i + \tilde{\eta}^T\right]\,,
\end{equation}
where the transformed parameters are defined as follows:
$W'_{ii} = \frac{2}{\sqrt{\epsilon}}$,
$W'_{ij} = \frac{\sqrt{\epsilon}}{2\lambda_\epsilon}W_{ij}$, and
$b'_i = \left(\frac{\sqrt{\epsilon}}{2\lambda_{\epsilon}}b_i-
  \frac1{\sqrt{\epsilon}}\right)$.  Figure~\ref{fig:machines}
illustrates a comparison between the structure of the Boltzmann
machine and the update dynamics. The activation function of the
$\textnormal{LM}^2$ is given in the limit of $\epsilon\to0$ by a
logistic distribution,
\begin{equation}
\label{eq:activationfunction}
\lim\limits_{\epsilon\to0}P_{\textnormal{LM}^2}(z_i=1) = \frac{1}{1+
  \exp\left[-\sum_{j}W_{ij} z_j - b_i\right]}\,.
\end{equation}
The term \textit{Langevin machine} is chosen because of the similarity
of the network to the Boltzmann machine and to Langevin dynamics. The
adjective \textit{discrete} is added to avoid confusion with the
Langevin machine presented in~\cite{Neelakanta1991}. The noise term in
the dynamics can be chosen according to
equation~(\ref{eq:truncatednoise}), i.e., it can be a Gaussian noise
or a truncated Gaussian noise. The self-interaction term
$W'_{ii}\in\lbrace0, 2/\sqrt{\epsilon}\rbrace$ and the contribution
$-1/\sqrt{\epsilon}$ of the bias $b'_i$ lead in dependency of the
state of the neuron for small values of $\epsilon$ to a strong shift
of the mean value into a positive or negative direction. Respectively,
the neuron stays very long in an active regime or in an inactive
regime in the case of Gaussian noise. The process fluctuates between
two different fundamental descriptions. The addition
\textit{sign-dependent} is used to emphasize this property and to
point out that the so far presented dynamics is a particular
realisation of the~\textit{discrete Langevin machine}, a larger class
of network implementations with a Gaussian noise distribution. This is
discussed in more detail in Section~\ref{sec:wholepicture}. The
exponent "$2$" in the abbreviation signifies the fluctuation between
the two regimes. The implicit dynamics~(\ref{eq:langevinmachine})
allows different interpretations and implementations.

After an absorption of the Gaussian noise term into the bias, the resulting network
has a rectangular decision function and can be interpreted as a neural
network with a noisy bias. The
simplicity of the update rule might be especially for neuromorphic
systems very helpful for a computation of physical statistical
systems, which are Boltzmann distributed. The implementation of an
exponential function in the system is much more challenging than
generating Gaussian noise. The rectangular decision function further
coincides with the threshold function of spiking neurons. Accordingly, a
possible adaptation of the dynamics on neuromorphic systems is 
obtained by the introduction of an additional timescale and staggered
Gaussian noise peaks. In the present work we pursue an alternative approach which is discussed in the next section.

Instead of
performing an implicit update, it is also possible to explicitly
compute the probability for an activation of the neuron in the next
step. This probability is given by
\begin{equation}
W_{\textnormal{LM}^2}(z_i\to 1)=\Phi\left(W'_{ii}z_i+\sum_{j} W'_{ij}z_j + b'_i\right)\,.
\end{equation}
In contrast to the Boltzmann machine, the transition probability is
not the same probability as the activation
function~(\ref{eq:activationfunction}).

Finally, the $\textnormal{LM}^2$ exhibits a totally different dynamics
than the Boltzmann machine. The dynamics is characterised by a Gaussian noise term as stochastic input, a
self-interacting term, its simplicity and multiple possible
implementations. Transition probabilities and correlation times can be
easily controlled by usage of truncated Gaussian noise. Finite values
of $\epsilon$ lead to a greater or lesser extent to deviating
observables, depending on the structure of the Boltzmann machine. The
source of error is given by the error term of order
$\mathcal{O}\left(\epsilon m_i^3\right)$ in the Taylor expansion of the detailed balance equation. Here, $m_i$
corresponds to the total input for a neuron $i$, according to
Appendix~\ref{sec:derivationlangevinmachine}. Exact results of a
Boltzmann machine can be obtained by an extrapolation to the limit
$\epsilon$.

\section{Neuromorphic Hardware System}
\label{sec:neuromorphichardware}

\begin{figure}[t]
	\centering
	\includegraphics[width=\linewidth]{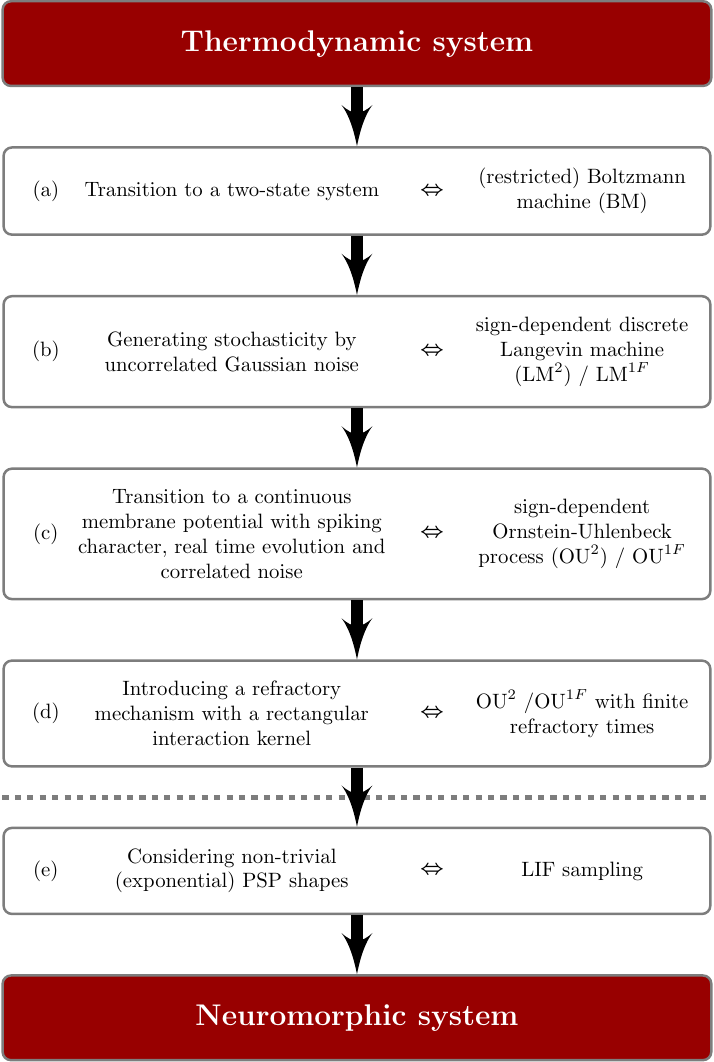}
	\caption{Illustration of a step-by-step approach to map
		thermodynamic systems on a neuromorphic hardware system. The
		dashed line indicates the progress of the paper. The paper
		proposes dynamics which have the potential to exactly
		preserve the properties of the Boltzmann machine up to this
		line.}
	\label{fig:flowdiagram}
\end{figure}

In this section, we discuss different approaches for a projection and
an accurate computation of the Boltzmann machine on a neuromorphic
hardware system. This is one of the possible application of the
sign-dependent Langevin machine. Several steps are necessary for a
successful projection, as indicated in
Figure~\ref{fig:flowdiagram}. Each step describes a different level of
abstraction of a neuromorphic system. A separate consideration of
different aspects of such a system enables a clear distinction and
identification of different sources of errors. Note that the diagram
in Figure~\ref{fig:flowdiagram} is not the only possible approach to 
such a projection.

In~\cite{Petrovici2016}, an analytic expression for the neural
activation function of leaky integrate-and-fire (LIF) neurons has been
derived for the hardware of the BrainScaleS project in Heidelberg. It
was demonstrated how the neuromorphic hardware system can be used to
perform stochastic inference with spiking neurons in the
high-conductance state. The microscopic dynamics of the membrane
potential of a neuron can be approximated in this state with
Poisson-driven LIF neurons by an Ornstein-Uhlenbeck process. The
spiking character of the system is obtained by a threshold function
which maps the system onto an effective two-state system.

Particular properties of LIF sampling are the following:
\begin{itemize} 
\item[(1)] A description of the
microscopic state of a neuron by a continuous membrane potential,
\item[(2)] An autocorrelated noise contribution to the membrane potential,
\item[(3)] A spiking character with an asymmetric refractory mechanism and
\item[(4)] Nontrivial and nonconstant interaction kernels between
  neurons.
\end{itemize}
In the present section we study simplified dynamics of LIF
sampling. This allows us to analyze the impact of particular hardware
related properties and sources of resulting errors on different levels
of abstraction. After a short introduction to the principles of LIF
sampling, a mapping of the $\textnormal{LM}^2$ is presented with
respect to several particularities of the hardware system. Possible
sources of errors of the mapping are discussed. We also relate our approach to the standard approach which relies on a fit of the
activation function of the hardware system to the activation function
of the Boltzmann machine.

The section ends with an analysis of the impact of a refractory
mechanism on the dynamics as a further step towards LIF sampling.

\subsection{LIF sampling}

The spikey neuromorphic system of the BrainScaleS project emulates
spiking neural networks with physical models of neurons and synapses
implemented in mixed-signal
microelectronics~\cite{Schemmel,Petrovici2016}. With the help of Poisson-driven leaky integrate-and-fire (LIF)
neurons, it is possible to obtain stochastic inference with
deterministic spiking neurons. The dynamics of the free membrane
potential $u_\textnormal{eff}(t)$ of a neuron can be approximated in the
high-conductance state by an Ornstein-Uhlenbeck process, 
\begin{equation}
\label{eq:closedform}
\frac{du_\textnormal{eff}(t)}{dt} = \theta\left[\mu-u_\textnormal{eff}(t)
\right]+\sigma \tilde{\eta}(t)\,. 
\end{equation}
In \eq{eq:closedform}, $\theta$ determines the strength of the
attractive force towards the mean value
$\mu=\mu^\textnormal{leak}+\mu^\textnormal{average noise}$. The mean
value consists of some leak potential and an additional averaged noise
contribution. The parameter $\sigma$ depends on the contribution of the Poisson
background.

\begin{figure}[htbp]
	\centering\includegraphics[width = 1.0\linewidth]{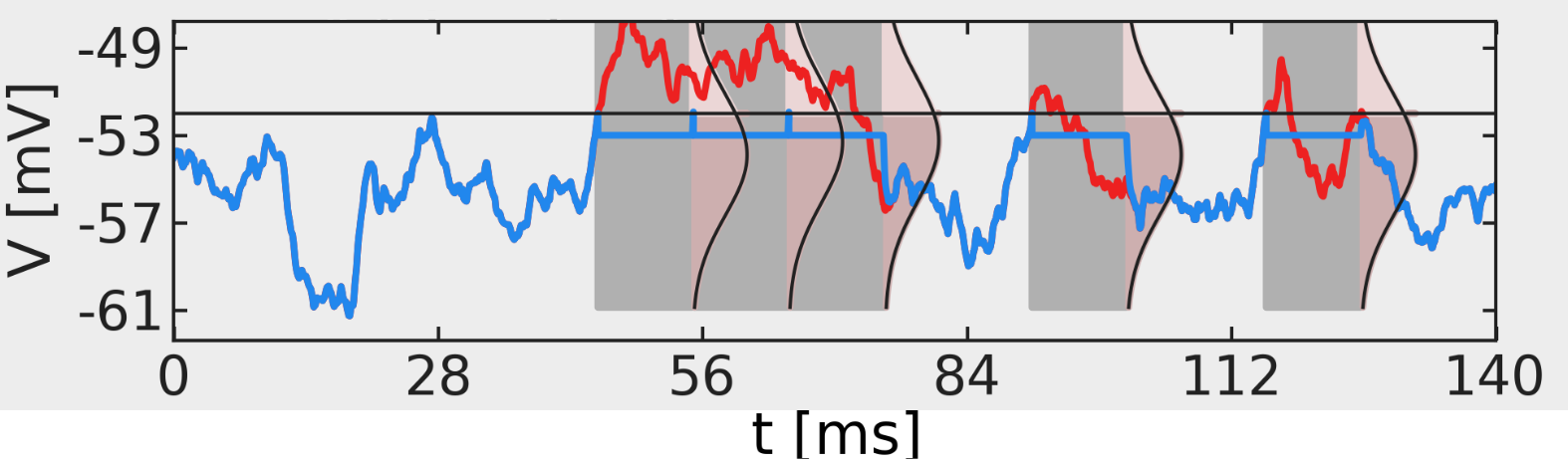}
	\caption[Example evolution of the free and the actual membrane
        potential]{Example evolution of the free membrane potential
          (red) and the actual membrane potential (blue). After the
          membrane potential crosses the threshold $\vartheta$, a
          spike is emitted and the potential is set to a reset
          potential $\varrho$. If the effective membrane potential is
          still above $\vartheta$ after the refractory period
          $\tau_\textnormal{ref}$, the neuron spikes again. At the end
          of a "burst" of $n$ spikes the free and the actual membrane
          potential converge in negligible time (scheme taken
          from~\cite{Petrovici2016b}).}
	\label{fig:membraneevolution}
\end{figure}

Inspired by a biological neuron~\cite{Petrovici2016b}, the neuron
emits a spike when the membrane potential exceeds a certain threshold
$\vartheta$. It is active and is reset to $\varrho$ for a refractory
time $\tau_\textnormal{ref}$ afterwards, where the neuron is
considered as inactive. This is also sketched in
Figure~\ref{fig:membraneevolution}. One has to distinguish between the
effective membrane potential $u_\textnormal{eff}(t)$ (red curve),
which is unaffected by the spiking dynamics, and the real membrane
potential $u(t)$ (blue curve). As in~\cite{Petrovici2016}, it is
assumed that the convergence of $u(t)$ from $\varrho$ to
$u_\textnormal{eff}(t)$ takes place in a negligible time after the
finite refractory time has elapsed.

\subsubsection{Activation function}

One can calculate distributions for the so-called burst lengths and
the mean first passage times of the membrane potential with the help
of transition probabilities $p(u_{i+1}|u_i)$. These are given by a
corresponding Fokker-Planck equation of the Ornstein-Uhlenbeck process
in the high conductance state. The burst length $n$ is the number of consecutive spikes. The mean
first passage time corresponds to the mean duration which it takes for
the membrane potential to pass the threshold $\vartheta$ from a lower
starting point. This is the time between an end of a burst and the
next spike. From an iterative calculation one can derive an activation
function, i.e., a probability distribution for the neuron to be active
($z=1$), in terms of these probabilities~\cite{Petrovici2016}, 
\begin{equation}
\label{eq:activationLIFneuron}
P(z=1)=\frac{\sum_n P_n n \tau_\textnormal{ref}}{\sum_n P_n
  \left(n\tau_\textnormal{ref} + \sum_{k=1}^{n-1} \overline{\tau_k^b} + T_n\right)}\,,
\end{equation}
where $\overline{\tau_k^b}$ corresponds to the mean drift time from
the resting potential $\varrho$ to $\vartheta$. The distribution over
burst lengths is represented by $P_n$, and the distribution over the
mean times between burst regimes is given by $T_n$.

\subsubsection{A simplified model}

If the refractory time $\tau_\textnormal{ref}$ is neglected, the
neuron can be interpreted as active if the effective membrane
potential is above a certain threshold, and as inactive otherwise. The
resulting dynamics corresponds to the level of abstraction~(c) in
Figure~\ref{fig:flowdiagram}. The neuron state is given by
$z(t):=\Theta\left[u_\textnormal{eff}(t)-\vartheta\right]$. The
process is referred to as an Ornstein-Uhlenbeck process with spiking
character. Interacting contributions are implemented on the basis of
the projected neuron states instead of their actual effective
continuous membrane potential. The activation function is a cumulative
Gaussian distribution, 
\begin{equation}
\label{eq:activationOrnStein}
P_\textnormal{eq}(z=1)=\int_{\vartheta}^{\infty}
P_\textnormal{eq}(u_\textnormal{eff})\textnormal{d}u_\textnormal{eff}=
\Phi\left(\frac{\sqrt{2\theta}}{\sigma}(\mu-\vartheta)\right)\,,
\end{equation}
with the equilibrium distribution $P_\textnormal{eq}$ of the Ornstein-Uhlenbeck
process~(\ref{eq:closedform}), 
\begin{equation}
\label{eq:activationUeff}
P_\textnormal{eq}(u_\textnormal{eff})=\sqrt{\frac{\theta}{\pi \sigma^2}}
\exp\left(-\frac{\theta(u_\textnormal{eff}-\mu)^2}{\sigma^2}\right)\,.
\end{equation}
For convenience, the threshold
potential $\vartheta$ is set to zero in further considerations.

The more accurate expression of the activation function, given by
equation~(\ref{eq:activationLIFneuron}), takes the finite refractory
time into account. The actual activation function is somewhere between
a logistic distribution and the cumulative Gaussian distribution.

In the following, we neglect the finiteness of the refractory
time. Therefore, we consider mostly simplified theoretical models of
the hardware system of the level of abstraction~(c). The models can be
used to analyze Gaussian noise as stochastic input and the impact
of autocorrelated noise in a system with a microscopic
real time evolution of the membrane potential. Given by the spiking
character, the considered models correspond effectively to interacting
two-state systems. A discussion with respect to a refractory
mechanism, as a process of the level of abstraction~(d), is given in
Section~\ref{sec:refractorymechanism}.

\subsection{Boltzmann machine}
\label{sec:mappingboltzmannmachine}

In this section we discuss a mapping of the Boltzmann machine (BM)
onto an Ornstein-Uhlenbeck process with spiking character as a
simplified model of a neuromorphic hardware system. A successful
mapping demands a logistic distribution as activation function and a
correct handling of interactions between neurons.

A possible approach is a fit of the activation function with a scaling
parameter $r$ and a shift parameter $\mu^0$ to the desired logistic
distribution according to
$P_\textnormal{eq}(z=1)=\Phi\left(\frac{\sqrt{2\theta}}{\sigma}
  \frac{\mu-\mu^0}{r}\right)\approx\sigma\left(\mu\right)$, with
$\sigma(x) = \left[1+\exp(-x)\right]^{-1}$~\cite{Petrovici2016, Leng,
  Probst}. Interactions can be taken into account by absorbing their
contributions into the mean value $\mu$ of the Ornstein-Uhlbenbeck
process according to: $\mu \to \mu+\mu^\textnormal{interaction}$. It
is assumed that the time to equilibrium is negligible after a change
of an interacting neuron.

A mapping of the Boltzmann machine to a process of the level of
abstraction~(c) can be performed straightforwardly by identifying the
total input (see Appendix~\ref{sec:derivationlangevinmachine}) of a
neuron $i$ with the mean value $\mu_i$ of the Ornstein-Uhlenbeck
process according to: $\mu_i \leftrightarrow -m_i$. This can be
achieved by setting the average noise contribution to zero, adjusting
the leak potential to $b_i$, and by taking the interacting
contributions into account:
\begin{align}
  \mu_i^\textnormal{average noise}&=0\,,\nonumber\\[1ex]
  \mu_i^\textnormal{leak}&=b_i\,,\nonumber\\[1ex]
  \mu_i^\textnormal{interaction}&=\sum_{\textnormal{syn} j} W_{ij} z_j(t)\,.
\end{align} 
Then, from the dynamics of equation~(\ref{eq:closedform}) with a
correct scaling of the interaction strength (see
Appendix~\ref{sec:interaction}) the following Ornstein-Uhlenbeck
process with spiking character is obtained: 
\begin{align}
\label{eq:ornsteinboltzmann}
  \frac{du_{i,\textnormal{eff}}(t)}{dt} &=\nonumber\\[1ex]
  =\frac{\theta}{r^2}&\left[\sum_{\textnormal{syn} j} W_{ij} z_j(t)+b_i
                       -\mu_i^0-u_{i,\textnormal{eff}}(t)\right]+\sigma
                       \tilde{\eta}(t)\,,
\end{align}
with $z_j(t)=\Theta\left[u_{j,\textnormal{eff}}(t)-\vartheta\right]$
and where $W_{ij}=\frac{A_{ij}}{\alpha}$ and
$A_{ij}\to \frac{A_{ij}}{\alpha}$. With $\sigma=\sqrt{2}$, $\theta=1$
we obtain the following activation function:
\begin{equation}
\label{eq:activationOrnSteinMem}
P_{\textnormal{OU}^{1\textnormal{F}}}(z_i=1)=\Phi\left(\frac{\sum_{\textnormal{syn} j} W_{ij} z_j + b_i - \mu_i^0}{r}\right)\,.
\end{equation}
The process is abbreviated in the following by
$\textnormal{OU}^{1\textnormal{F}}$. The "1" in the exponent is chosen
in compliance with the $\textnormal{LM}^2$ process and indicates that
the process takes place in one regime, i.e., the process does not
fluctuate between two fundamental dynamics. The fitting of the
activation function to the logistic distribution is indicated by the
additional "F". The process without any fitting parameters ($r=1$,
$\mu_i^0=0$) is denoted as $\textnormal{OU}^1$.

We can also formulate an update rule with the same resulting
activation function for a discrete two-state system, i.e., a system
without a membrane potential. The resulting system is build upon an
immediate representation of the neuron state, as it is the case for
the BM and the $\textnormal{LM}^2$. The resulting process corresponds
to a transition from the level of abstraction~(c) to the level (b) and
is driven by uncorrelated Gaussian noise.

The related update rule of level of abstraction~(b) is derived in a
similar manner as for the Langevin equation for discrete systems. It
is given by, 
\begin{equation}
\label{eq:ornsteinmachine}
z_i'=\Theta\left[\sum_{\textnormal{syn} j} W_{ij}z_j + b_i + \tilde\eta\right]\,,
\end{equation}
where the updates take place in computer time. The corresponding
transition probability reads, 
\begin{equation}
W_{\textnormal{LM}^{1\textnormal{F}}}(z_i\to1)=\Phi\left(\frac{\sum_{\textnormal{syn} j} W_{ij} z_j + b_i - \mu_i^0}{r}\right)\,.
\end{equation}
The dynamics has an additive Gaussian noise term and the Heaviside
function as a projection onto the domain of definition of
$z_i$. Therefore, it is very similar to the sign-dependent discrete
Langevin machine~(\ref{eq:langevinmachine}). The update rule is
studied in~\cite{Jordan2015} in more detail and introduced in~\cite{Emre2016}
as an approximation of the so-called \textit{Synaptic Sampling Machine}. In
compliance with the $\textnormal{LM}^2$ and the $\textnormal{OU}^1$,
we use the abbreviation $\textnormal{LM}^1$ for the process. When the
activation function is fitted to the logistic distribution,
$\textnormal{LM}^{1\textnormal{F}}$ is the corresponding acronym. In
the latter case, sources of errors are resulting deviations due to an
imperfect fit and finite times to equilibrium if an interacting neuron
changes its macroscopic state~\cite{Jordan2015}.

The dynamics can be interpreted as another realisation of the discrete
Langevin machine, as will be discussed in
Section~\ref{sec:wholepicture}. Properties and similarities of the two
presented processes, i.e., the
$\textnormal{LM}^1$~(\ref{eq:ornsteinmachine}) and the
$\textnormal{OU}^1$~(\ref{eq:closedform}) and
(\ref{eq:ornsteinboltzmann}), are numerically investigated in
Section~\ref{sec:applications}.

\subsection{Sign-dependent Ornstein-Uhlenbeck process}
\label{sec:mappinglangevinmachine}

The sign-dependent discrete Langevin machine and LIF neurons exhibit
similar underlying dynamics. This motivates a mapping of the
$\textnormal{LM}^2$ onto an Ornstein-Uhlenbeck process with spiking
character in the same manner as in the previous section, i.e., from the
level of abstraction~(b) to (c). The resulting process represents a
continuous counterpart to the sign-dependent discrete Langevin machine
and is referred to as \textit{sign-dependent Ornstein-Uhlbenbeck
  process} ($\textnormal{OU}^2$). The activation function of the
$\textnormal{OU}^2$ process converges in the limit $\epsilon\to 0$ also to a logistic
function. As illustrated in Figure~\ref{fig:systems}, the two
processes differ in their microscopic representation and their timescales. The $\textnormal{LM}^2$ corresponds to a process with two
discrete states and the computer time as timescale. The
$\textnormal{OU}^2$ process describes the temporal evolution of a
membrane potential in real time, whereas the interactions between
neurons are based on a projection of the potential onto two states.

\begin{figure*}[htbp]
  \centering\resizebox{0.49\linewidth}{!}{\includestandalone[mode=buildnew]{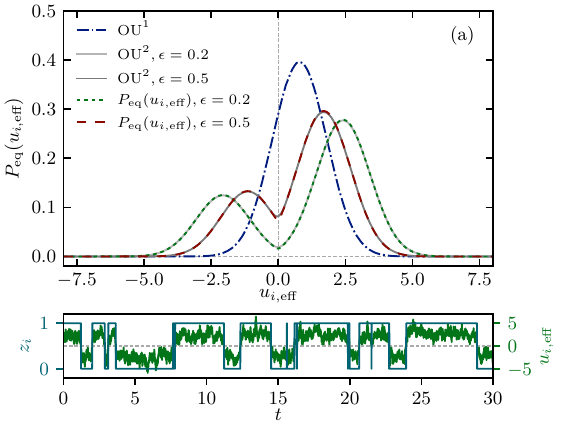}}
  \hfill\resizebox{0.49\linewidth}{!}{\includestandalone[mode=buildnew]{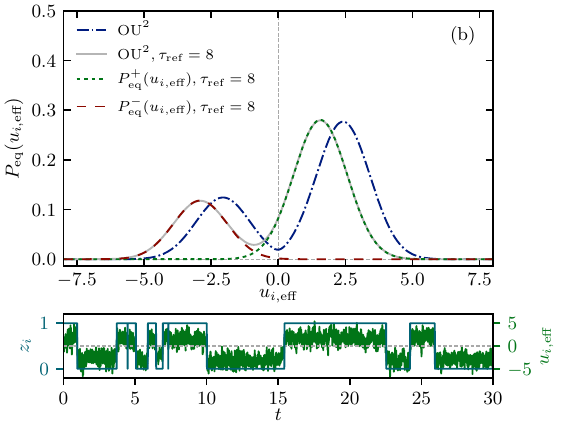}}
  \caption{Upper plots: Comparison of equilibrium distributions
    $P_\textnormal{eq}(u_{i, \textnormal{eff}})$ for the continuous
    processes for the free case with bias $b=0.8$ without a refractory mechanism~(a) and with a refractory mechanism with refractory time $\tau_{\textnormal{ref}}=8$~(b). Lower plots: Corresponding example trajectories of the
    $\textnormal{OU}^2$ processes with and without a refractory mechanism for $\epsilon=0.2$. The
    timescale of the lower plot is rescaled according to the
    transition probabilities to coincide.}
\label{fig:peqnref}
\end{figure*}

The total input $\sum_j W_{ij} z_j + b_i$ of the dynamics of the
previous section is exchanged for a mapping onto an Ornstein-Uhlenbeck
process by the redefined membrane potential of the sign-dependent
discrete Langevin machine: $W'_{ii}z_i+\sum_{j} W'_{ij}z_j +
b'_i$. This leads to the following dynamics of the sign-dependent
Ornstein-Uhlenbeck process:
\begin{align}
\label{eq:dynOU}
  \frac{du_{i,\textnormal{eff}}(t)}{dt} &=\nonumber\\[1ex] =
  \theta&\left[W'_{ii}z_i(t)+\sum_{\textnormal{syn} j} W'_{ij}z_j(t)
          + b'_i-u_{i,\textnormal{eff}}(t)\right]\nonumber\\[1ex]
                                        &+\sigma \tilde{\eta}(t)\,,
\end{align}
with
$z_i(t)=\Theta\left[u_{i,\textnormal{eff}}(t)-\vartheta\right]$. The
additional scaling factor of $\lambda_\epsilon$ is omitted, i.e., it
holds: $W_{ii}'=\frac{2}{\sqrt{\epsilon}}$,
$W_{ij}'=\frac{\sqrt{\epsilon}}{2} W_{ij}$, and
$b_{i}'=\left(\frac{\sqrt{\epsilon}}{2}
  b_{i}-\frac{1}{\sqrt{\epsilon}}\right)$.  The term
\textit{sign-dependent} reflects again the property of the neuron to
stay very long in an active regime ("$+$"), or in an
inactive regime ("$-$"). If the membrane potential randomly crosses
	the threshold $\vartheta$, it perceives a strong drift towards the
	other regime, due to the changing mean value of the process, i.e., due to the transition
	$z_i=0 \to z_i=1$. An immediate return to its initial regime gets rather
	unlikely because of the dependence of the mean value on $\frac{1}{\sqrt{\epsilon}}$. This property is reflected by the characteristic dynamics of the process which is shown in the lower left part of Figure~\ref{fig:peqnref}.

The equilibrium distribution over $u_{i, \textnormal{eff}}(t)$ of the process is derived in Appendix~\ref{sec:statsigndependent}. It is given by
\begin{equation}
\label{eq:eqOU}
P_\textnormal{eq}(u_{i,\textnormal{eff}}) \propto \exp\left[K(u_{i,\textnormal{eff}})\right]\,,
\end{equation}
with
\begin{align}
K&(u_{i,\textnormal{eff}})=\nonumber\\[1ex]
&=\frac{|u_{i,\textnormal{eff}}|}{\sqrt{\epsilon}}+\frac{\sqrt{\epsilon}}{2}\left(\sum_{\textnormal{syn} j} W_{ij}z_j
+ b_i\right)u_{i,\textnormal{eff}}-\frac{u_{i,\textnormal{eff}}^2}{2}\,.
\end{align}
The exponent $K(u_{i,\textnormal{eff}})$ contains the absolute value of the effective potential. This property causes a change of the sign of the resulting Gaussian distribution in dependence of the sign of $u_{i, \textnormal{eff}}$. Therefore, the equilibrium distribution can be considered as the concatenation of two Gaussian distributions $P_\textnormal{eq}^+(u_{i,\textnormal{eff}})$ and
$P_\textnormal{eq}^-(u_{i,\textnormal{eff}})$ with different mean values. In the active regime, the resulting mean value is $\mu_i^+(t)=-\frac{1}{\sqrt{\epsilon}} - \frac{\sqrt{\epsilon}}{2}
\left(\sum_{\textnormal{syn} j} W_{ij}z_j(t) + b_i\right)$ and in the inactive regime, it holds $\mu_i^-(t)=-\frac{1}{\sqrt{\epsilon}} + \frac{\sqrt{\epsilon}}{2}
\left(\sum_{\textnormal{syn} j} W_{ij}z_j(t) + b_i\right)$. The two distribution are reweighted according to the resulting overall stationary probability distribution. The left part of Figure~\ref{fig:peqnref} compares numerically found stationary distributions with the analytical result of equation~(\ref{eq:eqOU}).

\begin{figure*}[htbp]
	\centering\resizebox{0.47\linewidth}{!}{\includestandalone[mode=buildnew]{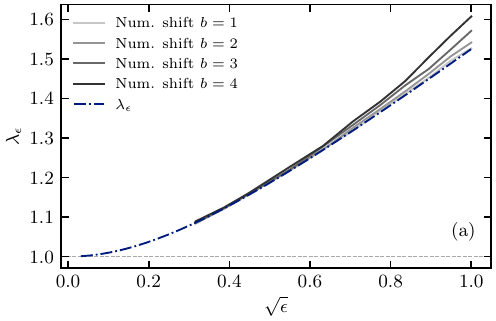}}
  \hfill\resizebox{0.47\linewidth}{!}{\includestandalone[mode=buildnew]{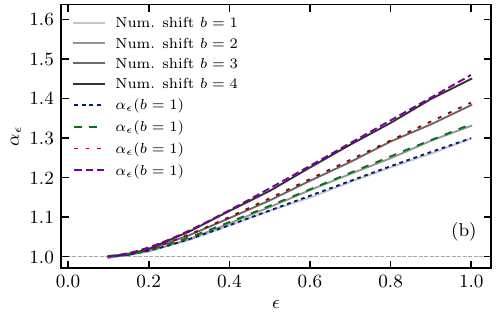}}
  \caption{Comparison of scaling and correction factors for the sign-dependent
    processes for a better convergence to the logistic
    distribution: (a)~The numerical and the analytic scaling factor
    $\lambda_{\epsilon} = \lambda\left(\sqrt{\epsilon}\right)$ for the
    sign-dependent Langevin
    machine. (b)~The numerical and analytic correction factors $\alpha_{\epsilon}$ for
    the sign-dependent Ornstein-Uhlenbeck
    process.}
	\label{fig:reps}
\end{figure*}

The stationary distribution of the process as a function of $z_i$ can be obtained by an integration of $P_\textnormal{eq}(u_{i,\textnormal{eff}})$ over $u_{i,\textnormal{eff}}$ with respect ot the threshold $\vartheta$. As derived in Appendix~\ref{sec:statsigndependent}, this results in the following activation function of the sign-dependent Ornstein-Uhlenbeck process,
\begin{equation}
P_{\textnormal{OU}^2}(z_i=1)=\frac{1}{1+\exp\left[\alpha_{\epsilon}(m_i)\times m_i\right]}\,,
\end{equation}
with $\alpha_{\epsilon}(m_i)$ being a correction factor. Here, $m_i:=-\sum_{\textnormal{syn} j} W_{ij}z_j(t) - b_i$ corresponds to the total input of the neuron. Since $\lim\limits_{\epsilon\to 0} \alpha_{\epsilon}(m_i)=1$, the activation function converges in the limit of $\epsilon\to 0$ to the logistic distribution,
\begin{equation}
\lim\limits_{\epsilon\to 0}P_{\textnormal{OU}^2}(z_i=1)=\frac{1}{1+\exp\left[-
	\sum_{\textnormal{syn} j}W_{ij} z_j - b_i\right]}\,,
\end{equation}
the activation function of the Boltzmann machine. The correction factor $\alpha_{\epsilon}(m_i)$ is different to the scaling factor
$\lambda_\epsilon$ of the $\textnormal{LM}^2$ since the convergence to the logistic distribution is caused by different characteristics for the two processes. Figure~\ref{fig:reps} compares numerically found scaling and correction
factors of the $\textnormal{LM}^2$ and $\textnormal{OU}^2$ process with the theoretical factors $\lambda_\epsilon$ and $\alpha_{\epsilon}(m_i)$ for
different biases $b$ in the free case, i.e., for the activation
function.

The overlap of the tails of the two shifted Gaussian distributions is responsible for the correction factor. The two distributions do not overlap at all in the limit of $\epsilon \to 0$. Hence, the activation function corresponds to the logistic distribution only in this case. For larger values of $\epsilon$, the distributions are closer together and crossings between the active and the inactive state take place more often. This property results in deviations to the logistic distribution.

An advantage of the sign-dependent Ornstein-Uhlenbeck process is the
possibility to extrapolate results for different values of $\epsilon$
to the limit of $\epsilon\to0$, i.e., to exact results of the
Boltzmann machine. A disadvantage of the process is that smaller values of
$\epsilon$ lead to larger correlation times and therefore to a higher
simulation cost. This results from the limitation that it is not
possible to accelerate the dynamics by an adaptation of the noise
source, as it is the case for the $\textnormal{LM}^2$. From another
perspective, this property might even help to straighten out problems
related to the hardware, like nontrivial postsynaptic shapes, for
example.

A comparison of the dynamics in~(\ref{eq:dynOU}) with the mapping of
the Boltzmann machine in equation~(\ref{eq:ornsteinboltzmann}) shows
that the self-interaction and the dynamics of the discrete Langevin
machine are key ingredients for a successful mapping onto the spiking
system.  The properties of the $\textnormal{OU}^2$ process are
dominated essentially by the self-interacting term. Therefore, the
process is not just an Ornstein-Uhlenbeck process, but represents a kind of dynamics with a different resulting equilibrium
distribution and, up to now, noninvestigated properties. Similar
dynamics which contain projected values of interacting potentials
might serve as a starting point for an entire class of dynamics.

\subsection{Refractory mechanism}
\label{sec:refractorymechanism}

A possible further step towards LIF sampling is to take into account a
refractory mechanism. This step is given in
Figure~\ref{fig:flowdiagram} by the level of abstraction~(d). The
refractory mechanism can be also considered for a discrete system, for example, for the Boltzmann
machine. This approach represents a different ordering of the
different abstractions of Figure~\ref{fig:flowdiagram}.

In a simplified model, it can be assumed that a neuron stays active
for the refractory time $\tau_\textnormal{ref}$, after it got
activated. An imbalance between the active and the inactive state is
caused by this property. This asymmetry can be compensated by reducing
the transition probability to become active by a factor of
$1/\tau_{\textnormal{ref}}$, as discussed in~\cite{Buesing2011}. The
factor can be absorbed into the membrane potential by a shift of the
activation function by $\log\left(\tau_\textnormal{ref}\right)$, i.e.,
by $b_i\to b_i-\log\left(\tau_\textnormal{ref}\right)$. Note that the
sign-dependent processes lead to a reformulation of the \textit{neuron
  computability condition} of~\cite{Buesing2011} due to the inherent
dependency of the dynamics on the neuron state itself.

\begin{figure}[b]
	\includestandalone[mode=buildnew]{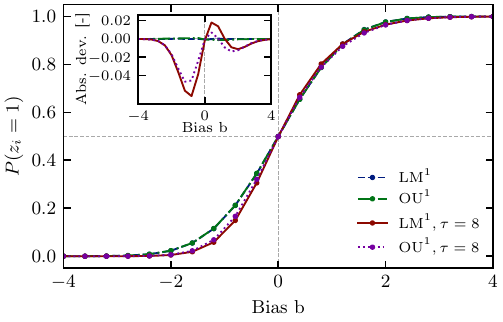}
	\caption{Illustration of the nonsymmetric deformation of the
          $\textnormal{LM}^1$ and the $\textnormal{OU}^1$ process for
          larger refractory times. The activation functions are
          shifted to comply: $P(z_i=1)\big|_{b=0}=0.5$. The small plot
          contains the absolute deviation to the cumulative Gaussian
          distribution.}
	\label{fig:cumulativedistributionrefractory}
\end{figure}

For the cumulative Gaussian distribution, an absorption of the factor
of $1/\tau_{\textnormal{ref}}$ is not possible anymore. The resulting
activation function with a finite refractory time is deformed. The
deformation gets larger for larger refractory times, as can be seen in
Figure~\ref{fig:cumulativedistributionrefractory}. We conclude that
the errors of the activation function to the logistic distribution
without a refractory mechanism propagate and increase for dynamics
with finite refractory times $\tau_\textnormal{ref}$. The resulting
deformation of the activation function can be identified as a further
source of error.

Within the last level of abstraction of Figure~\ref{fig:flowdiagram},
interactions between neurons or with the neuron itself are in general
not constant. The so-called postsynaptic potential (PSP) corresponds
to the received input potential of an interacting
neuron~\cite{Petrovici2016}. In Appendix~\ref{sec:interaction}, the
relation between a correct implementation of the weights based on the
interaction kernel is discussed in more detail. In this work, only
rectangular PSP shapes are considered. An investigation of exponential
PSP shapes is postponed to future work.

It is important to distinguish between the refractory mechanism as a
property of each neuron itself and the postsynaptic potential. The
latter needs only to be taken into account if interactions between
neurons are considered. In particular, this means that the PSP shape
affects only the activation function of the sign-dependent processes
due to their self-interacting contribution.

\begin{table*}
	\renewcommand{\arraystretch}{1.3}
	\small
	\begin{tabular}{p{2.9cm} p {1.6cm}p{1.6cm}p{1.6cm}p{1.6cm}p{1.6cm}p{1.6cm}p{1.6cm}} \toprule
		& \textbf{Gibbs sampling (BM)} & $\textnormal{\textbf{LM}}^{\textnormal{\textbf{1F}}}$ & $\textnormal{\textbf{LM}}^{\textnormal{\textbf{2}}}$ & $\textnormal{\textbf{OU}}^{\textnormal{\textbf{1F}}}$ & $\textnormal{\textbf{OU}}^{\textnormal{\textbf{2}}}$ & $\textnormal{\textbf{LM}}^{\textnormal{\textbf{1}}}$ & $\textnormal{\textbf{OU}}^{\textnormal{\textbf{1}}}$ \\ \midrule\midrule
		\textbf{Activation\newline function} & Logistic distribution & $\approx$ Logistic distribution& $\approx$ Logistic distribution & $\approx$ Logistic distribution& $\approx$ Logistic distribution & Cumulative Gaussian distribution & Cumulative Gaussian distribution \\
		\textbf{Microscopic\newline representation} & Discrete & Discrete & Discrete & Continuous & Continuous & Discrete & Continuous \\
		\textbf{Timescale} & Computer time & Computer time & Computer time & Real time & Real time & Computer time & Real time \\
		\midrule
		\textbf{Deviations\newline (free case)} & Exact & Small & Small & Small & Small & Exact & Exact\\
		\textbf{Extrapolation to exact solution?} & - & No & Yes & No & Yes & - & - \\
		\midrule
		\textbf{Deviations\newline (interacting case)} & Exact & Medium & Small & Large & Small & Exact & Exact\\
		\textbf{Extrapolation to exact solution?} & - & No & Yes & No & Yes & - & - \\
		\midrule
		\textbf{Control of a refractory mechanism} & Exact~\cite{Buesing2011} & Constant shift $\tau(\tau')$ & Constant shift $\tau(\tau')$ & Constant shift $\tau(\tau')$ & Constant shift $\tau(\tau')$ & Nontrivial shift $\tau(\tau', W_{ij}, b_i)$ & Nontrivial shift $\tau(\tau', W_{ij}, b_i)$\\
		\bottomrule
	\end{tabular}
	\caption{Comparison of the different analysed dynamics. An extrapolation to the exact solution and, therefore, a control of sources of errors is possible for the $\textnormal{LM}^2$ and the $\textnormal{OU}^2$ for both with and without a refractory mechanism.}
	\label{tab:overview}
\end{table*}

\section{Discrete Langevin machine}
\label{sec:wholepicture}

The Ornstein-Uhlenbeck process with a spiking dynamics and correlated
noise offers the possibility to simulate a discrete two-state system
by an underlying continuous dynamics. The discrete Langevin machine
can be interpreted as a discrete counterpart to those spiking systems
with uncorrelated noise, as indicated in Figure~\ref{fig:systems}. It
has been shown, that it is possible to map different realisations of
simplified theoretical models of the neuromorphic hardware system onto
a two-state system (see Figure~\ref{fig:flowdiagram} as well as the
dynamics~(\ref{eq:ornsteinboltzmann})~$\leftrightarrow$~
(\ref{eq:ornsteinmachine})
and~(\ref{eq:dynOU})~$\leftrightarrow$~(\ref{eq:langevinmachine})). Denoting
the hardware as $\textnormal{HW}(p)$, with parameters
$p=\lbrace W_{ij}, b_i, z_i, \epsilon \rbrace$, and the discrete
Langevin machine as $\textnormal{LM}(h)$, with $h:=h(p)$, we can state
the important relation,

\begin{equation}
\label{eq:mapping}
\lbrace \textnormal{LM}(h)\rbrace = \lbrace \textnormal{HW}(p) \rbrace\,,
\end{equation}
i.e., there exists a discrete two-state system for each set of
parameters $p$ of the hardware which emulates the dynamics of the
spiking system in discrete space.

In terms of update dynamics this corresponds to the mapping of 
\begin{equation}
  \frac{du_{i,\textnormal{eff}}(t)}{dt} = \theta\left[\mu_i(p)-
    u_{i,\textnormal{eff}}(t)\right]+\sigma \tilde{\eta}(t)\,,
\end{equation}
onto a discrete dynamics 
\begin{equation}
z_i'=\Theta\left[\mu_i(h(p))+\tilde{\eta}\right]\,,
\end{equation}
for all realisations of $p$ and with
$z_i=\Theta\left[u_{i,\textnormal{eff}}(t)\right]$.

Relation~(\ref{eq:mapping}) and the formal introduction of the
discrete Langevin machine can be seen as a theoretical framework to
describe different possible implementations as well as several levels
of abstraction of LIF sampling in terms of processes in real time with
a continuous membrane potential and spiking character and processes in
computer time with discrete states. For an exact mapping $h(p)$, the
magnitudes of the sources of error have to be
matched. Table~\ref{tab:overview} gives an overview of the presented
theoretical models and their properties regarding different levels of
abstraction of a neuromorphic system.

\section{Applications}
\label{sec:applications}

\begin{figure*}[t]
	\centering\resizebox{0.48\linewidth}{!}{\includestandalone[mode=buildnew]{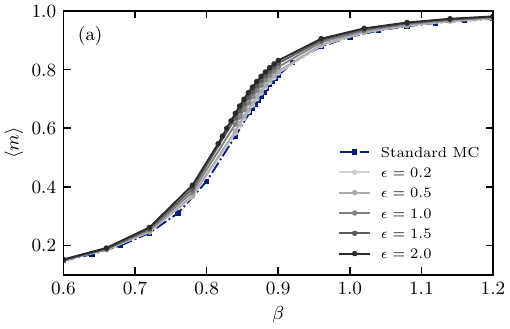}}
	\hfill\resizebox{0.48\linewidth}{!}{\includestandalone[mode=buildnew]{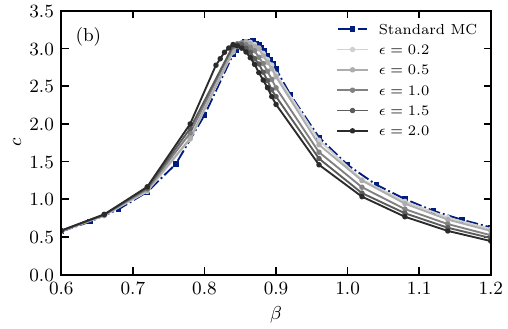}}
	\caption{Comparison of the magnetization~(a) and the specific heat~(b) obtained by a standard
		Monte Carlo algorithm and by the $\textnormal{LM}^2$ for
		the 4-state clock model on a $16\times16$ lattice. The
		results of the $\textnormal{LM}^2$ converge for
		$\epsilon\to0$ to the results of the standard Monte Carlo
		algorithm. Relative deviations of the inverse critical
		temperatures in dependence of $\epsilon$ are illustrated in
		Figure~\ref{fig:errorall}.}
	\label{fig:clockmodel}
\end{figure*}

Numerical results are discussed for the Langevin equation for discrete
systems of Section~\ref{sec:langevindynamicsdiscrete}, for the introduced sign-dependent Ornstein-Uhlenbeck process as well as for
existing approaches. We start with an analysis of the clock model in
Section~\ref{sec:clockmodel}. Dynamics and equilibrium
distributions of the free membrane potential are compared in
Sections~\ref{sec:neuromorphicvslangevin}
and~\ref{sec:refractorymechanismneuro} for the discrete Langevin
machine and for abstractions of the neuromorphic hardware system,
according to Figure~\ref{fig:flowdiagram} and
Table~\ref{tab:overview}. The focus is on a correct implementation of
the logistic distribution of the Boltzmann machine and on a detailed
analysis of the impact of different sources of errors. The systems are
considered with and without an asymmetric refractory mechanism with a
rectangular postsynaptic shape. The section ends with a computation
of the Ising model by a projection of the model on the Boltzmann
machine and with a numerical investigation of a Boltzmann machine with
three neurons in Section~\ref{sec:compbmising}. Both models serve
as a benchmark for Boltzmann distributed systems with interacting
neurons.

\begin{figure}[b]
	\centering{{\resizebox{0.96\linewidth}{!}{\includestandalone[mode=buildnew]{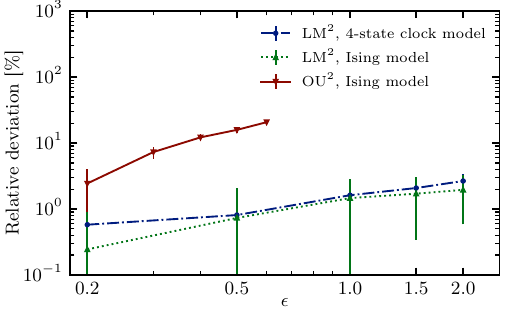}}}}
	\caption{Relative deviations of the obtained inverse critical
		temperatures for finite values of $\epsilon$ to the inverse
		critical temperature of a standard Monte Carlo algorithm.}
	\label{fig:errorall}
\end{figure}

\subsection{$q$-state clock model}
\label{sec:clockmodel}

The $q$-state clock model~\cite{Potts1951, Potts1952} describes spins
$\theta_i=\frac{2\pi n}q$ with $q$ different states which are
parametrised by $n\in\lbrace1,2,\dots,q\rbrace$. It is used to verify
numerically the Langevin equation for discrete systems, as a first
example. The model has the following Hamiltonian: 
\begin{equation}
\label{eq:Hamiltonianclock}
H_c = -J_c \sum_{\langle i,j\rangle} \cos\left(\theta_i-\theta_j\right)\,.
\end{equation}
The sum runs over all nearest neighbor spin pairs
$\langle i, j\rangle$. In a complex plane one can interpret the spin
states as equally distributed states on an unit circle. The common
Potts model~\cite{Wu1982} is derived from this initial model. For
$q=2$ the model corresponds to the Ising model and in the limit of
$q\to\infty$, it describes the continuous $XY$ model. For $q=4$ the
system emulates two independent Ising models.

The clock model exhibits for $q\leq4$ a second order phase
transition. It exists an exact solution for the inverse critical
temperature for $q=4$, which is as follows~\cite{Lapilli}:
\begin{equation}
J_c\beta_c^{q=4}=2J_c \beta_c^{q=2}\,,
\end{equation}
where the Boltzmann constant $k_b$ has been set to $1$. An appropriate
order parameter for the system is the average magnetization per spin,
which can be defined as 
\begin{equation}
\label{eq:magnetizationclock}
m = \frac1N \bigg|\sum_k^N e^{\frac{i2\pi n_k}{q}}\bigg|\,,
\end{equation}
where the sum runs over all spins $n_k$ of a lattice with $N$
sites. The specific heat capacity $c$ per spin is considered as a
further observable and is defined as 
\begin{equation}
c = \frac{\beta^2}{N}\left(\langle E^2 \rangle - \langle E \rangle^2 \right)\,,
\end{equation}
where $\langle \cdot \rangle$ denotes the expectation value~\cite{Newman1999}.

Numerical results for the magnetization and the specific heat are
illustrated in Figure~\ref{fig:clockmodel} in
dependency of the inverse temperature $\beta$ and of different values
of $\epsilon$. Results of the Metropolis algorithm as a standard Monte
Carlo algorithm (MC) serve as benchmark. The inverse critical
temperature can be read off from the maximum of the specific heat. In
Figure~\ref{fig:errorall}, the relative deviations of the inverse
critical temperature to the inverse critical temperature of the
Metropolis algorithm are plotted against $\epsilon$.

The resulting deviations for finite values of $\epsilon$ can be
explained by a detailed error analysis of the transition probabilities
of the Langevin equation for discrete systems. For this purpose, one
has to check the compliance of the detailed balance equation, 
\begin{equation}
  \frac{W_{\textnormal{LM}^2}(\theta\to \theta')}{W_{\textnormal{LM}^2}(
    \theta'\to \theta)}+\mathcal{O}\left(\epsilon \Delta H_c(\theta',
    \theta)^3\right) = \frac{P_{\textnormal{MC}}(\theta')}{P_{\textnormal{MC}}(\theta)}\,.
\end{equation}
In the left part of Figure~\ref{fig:convergenceexp}, it can be seen that the absolute error
of the cumulative Gaussian distribution is asymmetric around
$x=0$. This imbalance leads to a shift of the effective fraction of
transition probabilities and therefore to a change of the equilibrium
distribution of the spin states. The strength of this shift grows for
larger values of $x$, which corresponds to larger values of
$\beta\Delta H_c$, and larger values of $\epsilon$. The effect can be
nicely observed in the change of the specific heat in
the right part of Figure~\ref{fig:clockmodel} with growing $\beta$. In general, it holds:
the larger $|\beta \Delta H_c|$, the worse is the compliance of the
detailed balance equation and the larger is the resulting shift of the
equilibrium distribution.

\begin{figure*}[htbp]
	\centering\resizebox{0.48\textwidth}{!}{\includestandalone[mode=buildnew]{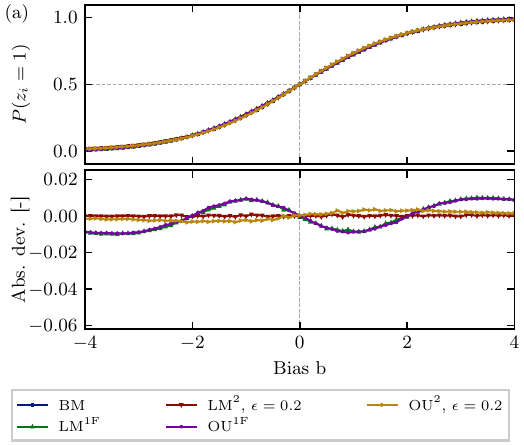}}\hfill
	\resizebox{0.48\textwidth}{!}{\includestandalone[mode=buildnew]{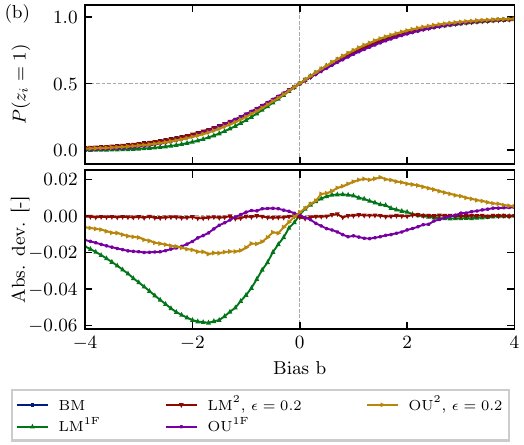}}\\\vspace{0.8cm}
	\centering\resizebox{0.48\linewidth}{!}{\includestandalone[mode=buildnew]{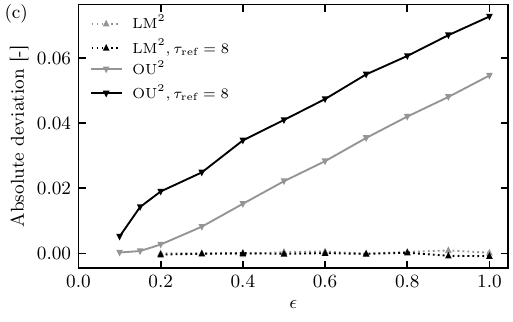}}
	\hfill\resizebox{0.48\linewidth}{!}{\includestandalone[mode=buildnew]{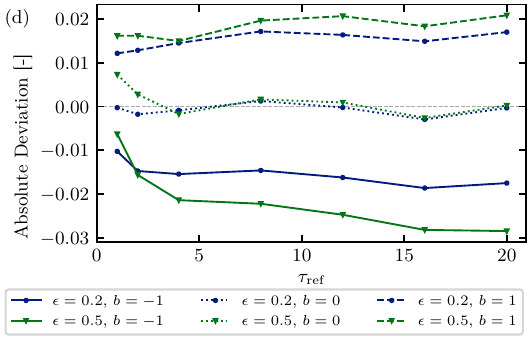}}
	\caption{Comparison of different properties of the activation
		function for the different processes with and without a refractory
		mechanism: (a)~Activation function and absolute deviation to
		the logistic distribution without a refractory
		mechanism. (b)~The same as in~(a), but with a refractory mechanism with refractory time $\tau=8$. (c)~Absolute deviation of the exact results of the
		activation for $b=1$ for the sign-dependent processes in dependence
		of $\epsilon$. The deviation converges in the limit $\epsilon\to0$
		for all processes to zero, with and without a refractory
		mechanism. (d)~Absolute deviation of the results for the
		$\textnormal{OU}^2$ process without a refractory mechanism to results with a
		refractory mechanism and different values of
		$\tau_\textnormal{ref}$. The deviations are compared for
		$b=\lbrace-1, 0,
		1\rbrace$.}
	\label{fig:convergencefdexp}
\end{figure*}

\subsection{Neuromorphic hardware versus Langevin machine}
\label{sec:neuromorphicvslangevin}

We analyze numerically the mapping between dynamics of the discrete
Langevin machine and the continuous dynamics according to
relation~(\ref{eq:mapping}) by an explicit consideration of transition
probabilities. It is discussed the impact of deviations in the
transition probabilities as well as a mapping of the temporal
evolution of the different processes onto each other with respect to
resulting activation functions for the free membrane potential.

Differences of the two dynamics which are given by construction are
illustrated in Figure~\ref{fig:systems}. The processes correspond to
the levels of abstraction (b) and (c) of a neurmorphic hardware system
in Figure~\ref{fig:flowdiagram}. The essential differences are the
source of noise, which is for the Langevin machine uncorrelated and for
the Ornstein-Uhlenbeck process correlated, as well as the
representation of a microscopic state. The dynamics is described in
the former case by two discrete states in computer time and in the
latter one by the evolution of a continuous membrane potential with
spiking character in real time. We evaluate the impact of the
different sources of errors on the deviation to the expected logistic
distribution for the sign-dependent and for the fitted processes:
$\textnormal{LM}^2$, $\textnormal{OU}^2$,
$\textnormal{LM}^{\textnormal{1F}}$, and
$\textnormal{OU}^{\textnormal{1F}}$.

\subsubsection{Activation function}

Figure~\ref{fig:cumulativedistributionrefractory}
and the upper left part of Figure~\ref{fig:convergencefdexp} illustrate the activation functions of the free
membrane potential in dependency of the bias $b$ in the network for
the different presented dynamics. The results of the
$\textnormal{LM}^1$ and the $\textnormal{OU}^1$ process coincide
exactly and their deviation to the cumulative Gaussian distribution
emerges from numerical errors. In concordance to these observations,
the fitted $\textnormal{LM}^{1\textnormal{F}}$ and
$\textnormal{OU}^{1\textnormal{F}}$ process have the same deviations
to the logistic distribution. In the case
of the $\textnormal{LM}^2$ and the $\textnormal{OU}^2$ process, the
observed deviations mirror the theoretical errors for finite values of
$\epsilon$. As depicted in the lower left part of Figure~\ref{fig:convergencefdexp}, both activation
functions converge in the limit of $\epsilon\to 0$. The rate of
convergence of the $\textnormal{OU}^2$ process is much smaller than
the one of the $\textnormal{LM}^2$ for equal values of
$\epsilon$. This can be reasoned by the different sources of errors
for the two processes, as discussed in detail at the ends of
Sections~\ref{sec:langevinmachine}
and~\ref{sec:mappinglangevinmachine}. In contrast to
the $\textnormal{LM}^{1\textnormal{F}}$ and
$\textnormal{OU}^{1\textnormal{F}}$, deviations to the logistic
distribution are limited only due to larger correlation times for
smaller values of $\epsilon$ for the $\textnormal{OU}^{2}$ process.

\subsubsection{Dynamics: time evolution}

\begin{figure}[t]
	\centering{\resizebox{1\linewidth}{!}{\includestandalone[mode=buildnew]{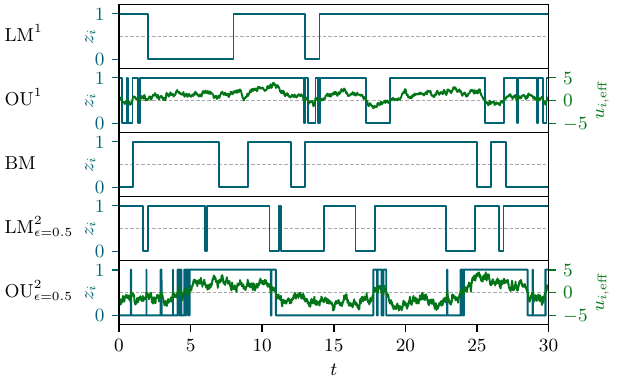}}}
	\caption{Trajectories of the neuron state and the membrane
          potential in computer time for the different processes with
          a uniform timescale.}
	\label{fig:timesdevolutdionsss}
\end{figure}

It has been found numerically that the computer time and the real time
coincide for the $\textnormal{LM}^1$ and the Ornstein-Uhlenbeck
process. All simulations in real time are performed with finite time
steps of $0.02$. All processes in computer time are computed with a
random sequential update formalism and in real time by a parallel
update scheme. The timescale in all figures is chosen in units of the
computer time.

Figure~\ref{fig:timesdevolutdionsss} compares trajectories of the
different discussed processes with respect to a uniform timescale. It
can be observed in the evolution of the membrane potential for all
processes that there occur fast changes if the membrane potential is
close to the threshold value $\vartheta=0$. These perturbations seem
to have no influence on the time evolution and the equilibrium
distribution.

\begin{figure}[t]
	\includestandalone[mode=buildnew]{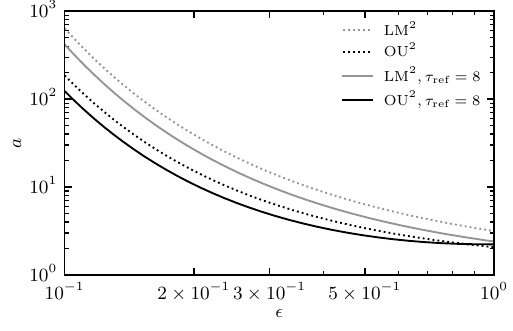}
	\caption{Scaling factors $a$ in dependence of $\epsilon$ for a
		mapping of the transition probabilities of the
		sign-dependent processes onto the transition probability of
		the Boltzmann machine and, hence, of the temporal evolution
		on the computer time. The scaling factors are computed for
		the free case with a bias $b=0$.}
	\label{fig:cumulativedistribdutionrefractory}
\end{figure}

As discussed in Section~\ref{sec:langevindynamicsdiscrete}, a scaling
factor $a$ can be found for a correct mapping of the temporal
evolution of two processes A and B if both processes exhibit the same
equilibrium distribution. The scaling factor is given under usage of
equation~(\ref{eq:rescalingrelation}) and by a computation of the
transition probabilities by 
\begin{equation}
a = \frac{W_{\textnormal{A}}(0\to 1)}{W_{\textnormal{B}}(0\to 1)}\,.
\end{equation}
Analytic expressions for the transition probabilities of the
considered sign-dependent processes are given in
Table~\ref{tab:transitionprobabilities}. The given transition
probabilities have been validated numerically. For that purpose we
have mapped the temporal ensemble evolution of the different dynamics
onto the evolution of the Boltzmann machine with respect to the
computed scaling factors. A scaling factor $a\neq1$ reflects the
increase/decrease of the correlation time for processes with different
transition probabilities. In
Figure~\ref{fig:cumulativedistribdutionrefractory}, the dependency of
the scaling factor $a$ on $\epsilon$ is plotted for the sign-dependent
processes.

The considerations of the time evolution reinforce that the relation
of equation~(\ref{eq:mapping}) corresponds to an exact mapping of the
dynamics of a discrete system with uncorrelated noise to a
continuous system with correlated noise. This property is not
self-evident. However, the dependency $h(p)$ is in some cases
nontrivial, due to different occurring sources of errors of the
considered models.

\subsection{Refractory mechanism}
\label{sec:refractorymechanismneuro}
In this section we investigate the impact of an asymmetric refractory
mechanism of a neuromorphic system with a rectangular PSP shape. This
has been introduced in Section~\ref{sec:refractorymechanism} as the
level of abstraction (d) with regard to Figure~\ref{fig:flowdiagram}.

\subsubsection{Control of the refractory mechanism}

We concentrate on a correct representation of the logistic function
with a refractory mechanism. The imbalance between the inactive and
the active state can in general not be compensated entirely by a
trivial shift of the bias by
$\log\left( \tau_{\textnormal{ref}}\right)$ for the cumulative
Gaussian distribution.  The activation functions of the sign-dependent
processes and the fitted dynamics are close to a logistic
distribution. Therefore, a shift can be used to approximate the
logistic distribution with a refractory mechanism. However, for large
refractory times this approximation gets worse, as indicated in
Figure~\ref{fig:cumulativedistributionrefractory}.

\begin{table*}
	\renewcommand{\arraystretch}{1.3} \small \begin{tabular}{cccc}
		\toprule & $W_{\textnormal{BM}}(0\to 1)$&
		$W_{\textnormal{LM}^2}(0\to 1)$& $W_{\textnormal{OU}^2}(0\to 1)$
		\\\midrule
		$\tau_{\textnormal{ref}}=1$ & $\sigma(-m_i)$ &  $\Phi\left(-\frac{1}{\sqrt{\epsilon}}-\frac{\sqrt{\epsilon}}{2\lambda_{\epsilon}}m_i\right)$& $\varphi\left(-\frac{1}{\sqrt{\epsilon}}-\frac{\sqrt{\epsilon}}{2}m_i\right)$ \\
		$\tau_{\textnormal{ref}}>1$
		&
		$\sigma(-\left(m_i+\log(\tau_{\textnormal{ref}})\right))$
		& 
		$\Phi\left(-\frac{1}{\sqrt{\epsilon}}-\frac{\sqrt{\epsilon}}{2\lambda_{\epsilon}}(m_i+\log(\tau_{\textnormal{ref}}))\right)$&
		$\varphi\left(-\frac{1}{\sqrt{\epsilon}}-\frac{\sqrt{\epsilon}}{2}(m_i+\log(\tau_{\textnormal{ref}}))\right)$
		\\\bottomrule
	\end{tabular}
	\caption{Transition probabilities from an inactive state to an
		active state for the different considered dynamics with
		($\tau_{\textnormal{ref}}>1$) and without
		($\tau_{\textnormal{ref}}=1$) a refractory time. $m_i$
		corresponds to the total input for a neuron $i$, according
		to
		Appendix~\ref{sec:derivationlangevinmachine}\label{tab:transitionprobabilities}.}
\end{table*}

In this work, the shift of the activation function is determined by
the constraint that $p(z_i = 1)\big|_{b=0} = 0.5$. The actual shifts
of the bias are slightly different for the
$\textnormal{LM}^{\textnormal{1F}}$ and the
$\textnormal{OU}^{\textnormal{1F}}$ as a consequence of resulting
deformations of the cumulative Gaussian distribution for larger
refractory times (see
Figure~\ref{fig:cumulativedistributionrefractory}). We also introduce
a further time constant $\tau_{\textnormal{ref}}'$. This allows us to
distinguish clearly between the refractory time
$\tau_\textnormal{ref}$ of a neuron and the resulting optimal shift
$\log(\tau_{\textnormal{ref}}')$ for a correct fixing of the
activation function. Ideally, one should derive a dependency
$\tau_{\textnormal{ref}}'(\tau_{\textnormal{ref}})$ to preserve a
consistent fixing of the activation function.

For the $\textnormal{LM}^2$ process it holds true that
$\tau_\textnormal{ref}\simeq\tau_\textnormal{ref}'$, since the
deviation of the activation function to the logistic distribution is
nearly symmetric around $b=0$. Nevertheless, a shift by $\log(\tau_\textnormal{ref}')$
leads to a worse approximation of the transition probability, since
the Taylor expansion of the cumulative Gaussian distribution of the
$\textnormal{LM}^2$ dynamics is performed around $0$. This is a
further error source.

\begin{figure}[htbp]
	{\resizebox{0.49\linewidth}{!}{\includestandalone[mode=buildnew]{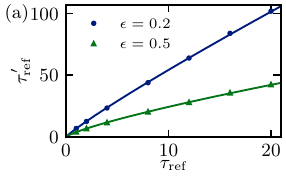}}}\hfill
	{\resizebox{0.50\linewidth}{!}{\includestandalone[mode=buildnew]{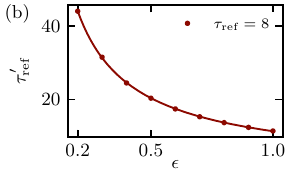}}}
	\caption{$\tau_{\textnormal{ref}}'$ in dependence of the
		refractory time $\tau_{\textnormal{ref}}$ for $\epsilon=0.2$
		and $\epsilon=0.5$~(a) and of $\epsilon$ for
		$\tau_{\textnormal{ref}}=8$~(b) for the
		$\textnormal{OU}^2$ process. Both dependencies obey a power
		law.}
	\label{fig:taurefvstaup}
\end{figure}

For the $\textnormal{OU}^{2}$ process, the necessary shift of the bias
by $\log\left(\tau_{\textnormal{ref}}'\right)$ is much larger than
$\log\left(\tau_{\textnormal{ref}}\right)$. Dependencies of
$\tau_\textnormal{ref}'(\tau_\textnormal{ref})$ for fixed values of
$\epsilon$ and $\tau_\textnormal{ref}'(\epsilon)$ for $\tau_\textnormal{ref}=8$ are
illustrated in Figure~\ref{fig:taurefvstaup}. The large differences in
$\tau_{\textnormal{ref}}$ and in $\tau_{\textnormal{ref}}'$ can be
traced back to the different microscopic dynamics of the processes and
to the different origin of a correct implementation of the activation
function for the processes without a refractory mechanism.

In contrast to the $\textnormal{OU}^1$ process, the dynamics of the
$\textnormal{OU}^2$ process fluctuates between the active and the
inactivate regime which are distinguished and driven by the
self-interacting contribution. Integrating a refractory mechanism, a
change from the active to the inactive regime by the self-interacting
term is suppressed as long as the neuron is captured in its refractory
mode. This can be seen in the lower right plot of
Figure~\ref{fig:peqnref}. Consequently, the two Gaussian distributions,
$P_\textnormal{eq}^+(u_{i,\textnormal{eff}})$ and
$P_\textnormal{eq}^-(u_{i,\textnormal{eff}})$, have a dissimilar impact
on the resulting distribution of $P(u_{i,\textnormal{eff}})$. The
lower tail distribution of $P_\textnormal{eq}^+(u_i)$, i.e., the part of the 
distribution for $u_{i,\textnormal{eff}} < \vartheta$, biases the
distribution $P(u_{i,\textnormal{eff}})$ around the threshold
$\vartheta=0$ within the refractory time. In contrast, the upper tail
distribution of $P_\textnormal{eq}^-(u_{i,\textnormal{eff}})$ does not
affect $P(u_{i,\textnormal{eff}})$, since the dynamics changes for
$u_{i,\textnormal{eff}}>\vartheta$ from the inactive to the active
regime.  The local minimum of $P(u_{i,\textnormal{eff}})$ around
$u_{i,\textnormal{eff}}=\vartheta$ as well as the entire distribution
$P(z_i)$ are shifted to smaller values, as a result of this asymmetry,
as illustrated in the upper right part of Figure~\ref{fig:peqnref}. Further, the absolute value
of the minimum is larger than the one for the process without a refractory
mechanism. The imbalance between
$P_\textnormal{eq}^+(u_{i,\textnormal{eff}})$ and
$P_\textnormal{eq}^-(u_{i,\textnormal{eff}})$ results in larger
deviations of the activation function for the $\textnormal{OU}^2$
process with a refractory mechanism. This asymmetry corresponds to a
further source of error.

The equilibrium distribution needs to undergo a larger shift by
$\log(\tau_\textnormal{ref}')$ than the
$\textnormal{OU}^{1\textnormal{F}}$ process for a compensation of the
impact of the refractory mechanism. This is a consequence of a
partially suppression of the change of the dynamics to the inactive
regime. For the $\textnormal{OU}^{1\textnormal{F}}$ process, the
underlying dynamics is not affected by the refractory mode due to the
absence of a self-interacting term. Therefore, the only purpose of the
shift by $\log\tau_\textnormal{ref}'$ is to fix the transition
probabilities to correctly compensate the emerging asymmetry of the
refractory mechanism. Respectively, the resulting transition
probabilities are expressed in
dependence of $\log(\tau_\textnormal{ref})$. Analytic expressions for the sign-dependent dynamics are
given in Table~\ref{tab:transitionprobabilities}.

\subsubsection{Activation function and time evolution}

The upper right part of Figure~\ref{fig:convergencefdexp} compares the impact of a refractory mechanism
on the different dynamics regarding their deviations to the logistic
distribution. The deviations of the $\textnormal{LM}^2$ and the
$\textnormal{OU}^2$ process have increased, as expected by the
introduced asymmetry of the refractory mechanism. Nevertheless, the
error vanishes for $\epsilon\to 0$, as illustrated in the lower left part of Figure~\ref{fig:convergencefdexp}. Further, the lower right part of Figure~\ref{fig:convergencefdexp} shows that a
further increase of the refractory time has a very low impact on the
deviations which ensures an applicability for large refractory times,
in practice. As discussed in Section~\ref{sec:refractorymechanism},
the cumulative Gaussian distribution is nonsymmetrically deformed by
the shift by $\log(\tau')$. This leads to
deviations in the activation function of the $\textnormal{LM}^1$ and
the $\textnormal{OU}^1$ process that can be compensated to a certain
extent by an adaptation of the variance, i.e., of the scaling parameter
$r$.

We conclude that a refractory mechanism with rectangular PSP shape has
no impact on a possible control of the sources of errors for the
sign-dependent processes.

\subsection{Interacting systems}
\label{sec:compbmising}

We consider the Ising model~\cite{Ising1925} and the Boltzmann
machine~\cite{Ackley} to investigate the presented abstractions of a
neuromorphic hardware system with interactions between neurons. The
Ising model can be easily mapped onto the Boltzmann machine. A
numerical analysis can be understood as a proof of concept that the
presented processes also work in a more complex network setup. As a
second model, we study a Boltzmann machine with three neurons. We
compare the results for all presented models with and without a
refractory mechanism with a rectangular PSP shape.

The Ising model describes a two-state spin system. The spin states are $s_i\in\lbrace -1,+1\rbrace$, which are
likewise also referred to as spin up and spin down
$s_i\in\lbrace\downarrow,\uparrow\rbrace$.  The Hamiltonian is defined
as 
\begin{equation}
H = -J\sum_{\langle i,j\rangle}s_i s_j - h \sum_i s_i\,.
\end{equation}
The external magnetic field $h$ is set to zero in the following
numerical analysis and $J=1$ is some coupling constant. For this
particular case, we can consider the averaged absolute value of the
magnetization per spin as an order parameter. This is then given by 
\begin{equation}
m = \frac1N\bigg|\sum_i^N s_i\bigg|\,,
\end{equation}
where the sum runs again over all spins of the lattice for a given
configuration. From theoretical considerations, an exact expression
for the inverse critical temperature of the model with a vanishing
external field can be obtained~\cite{Onsager1944}, 
\begin{equation}
J\beta_c=\frac{\ln\left(1+\sqrt{2}\right)}{2}\,.
\end{equation}
For a computation with the presented algorithms, we need a mapping
between the Boltzmann machine and the Ising model onto the correct
domain of definition. The mapping of $s_i=-1\to z_i=0$ and
$s_i=1\to z_i=1$ can be obtained by the following identifications
between $J$ and $h$ and $W_{ij}$ and $b_i$~\cite{Petrovici2016b,
  Bytschok2017SpikebasedPI}:
\begin{align}\nonumber 
W_{ij} &=  4 J\,, \\[1ex]
b_i &= 2 h - 2 J d\,,
\end{align}
where $d$ corresponds to the dimension of the system. The spin state
can be computed by $s_i = 2z_i-1$.

The Boltzmann machine can have an arbitrarily complex network
structure. Particular implementations like the restricted Boltzmann
machine turn the Boltzmann machine to an interesting class of
networks, which has many applications in different areas of
research; see, e.g.,~\cite{Czischek2018, Carleo2017, Gao2017, Montfar2018}. To
study the impact of systems with a higher possible variability, we
consider a Boltzmann machine with three neurons and different weights
and biases around zero. The Kullback-Leibler
divergence~\cite{kullback1951} serves as a measure to numerically
classify the quality of the presented processes. We compute the
Kullback-Leibler divergence based on the history of a process,
starting from a random initial state according to
\begin{equation}
  D_\textnormal{KL}\left(P_\textnormal{BM}||P_\textnormal{AM}\right) =
  -\sum_{c\in\Omega} P_\textnormal{BM}(c)
  \log \frac{P_\textnormal{AM}(c)}{P_\textnormal{BM}(c)}\,.
\end{equation}
BM indicates the exact probability distribution of the Boltzmann
machine and AM corresponds to the approximated probability
distribution of some other model. The sum runs over all possible
neuron configurations $c$. The probabilities are approximated by the
corresponding histograms of the history of the trajectory in the
configuration space.

\begin{figure*}[htbp]
	\centering\resizebox{0.48\textwidth}{!}{\includestandalone[mode=buildnew]{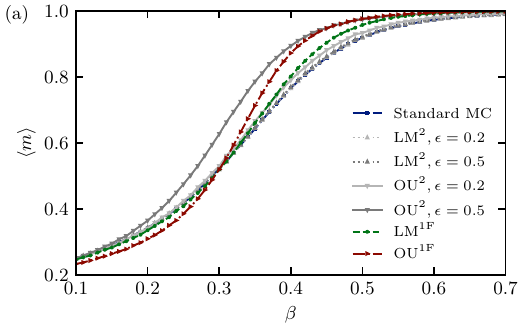}}\hfill\resizebox{0.48\textwidth}{!}{\includestandalone[mode=buildnew]{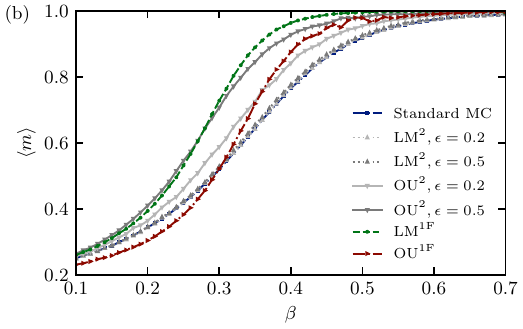}}\\\vspace{0.8cm}
	\centering\resizebox{0.48\textwidth}{!}{\includestandalone[mode=buildnew]{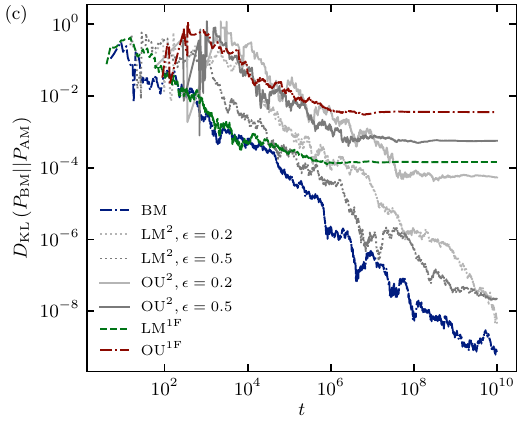}}
	\hfill\resizebox{0.48\textwidth}{!}{\includestandalone[mode=buildnew]{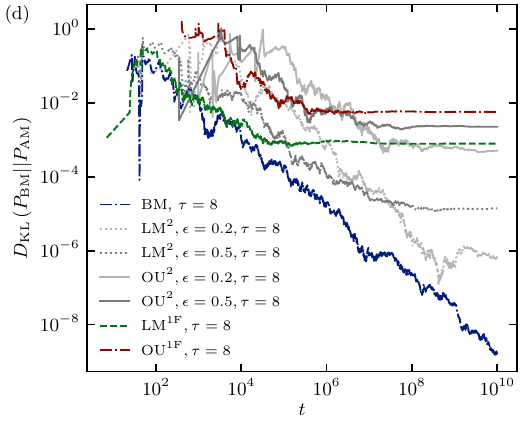}}
	\caption{Comparison of numerical results for the different
		models without a refractory mechanism (left column) and with
		a rectangular PSP shape: (a)~Ising model without a refractory mechanism ($4\times4$ lattice). (b)~Ising model with a rectangular PSP ($\tau=8$, $4\times4$ lattice). (c)~Kullback-Leibler divergence without a refractory
		mechanism. (d)~Kullback-Leibler divergence with a rectangular PSP
		($\tau=8$). (a),~(b): The absolute
		magnetization of the Ising model is plotted against the
		inverse temperature $\beta$. The deviations of the different
		models mirror the observed deviations of the activation
		function. The results confirm the presumption that small
		deviations in the activation function can have a large
		impact on the resulting observables. (c),~(d):
		Illustration of the evolution of the Kullback-Leibler
		divergence for a Boltzmann machine with three neurons based
		on their history. As an exception, the time is not rescaled
		with respect to the transition probabilities, i.e., the
		correlation times, in these plots. This causes a shift of
		the curves of the sign-dependent processes to larger
		times. The observed levels of convergence of the
		Kullback-Leibler divergence of the different models are in
		concordance to the results of the Ising model. The different
		levels of convergence for the fitted processes signifies the
		large dependency of the accuracy of resulting correlation
		functions on errors in the representation of the activation
		function and respectively on corresponding weights and
		biases within the network.}
	\label{fig:refractory}
\end{figure*}

\begin{figure}
	\resizebox{0.96\columnwidth}{!}{\includestandalone[mode=buildnew]{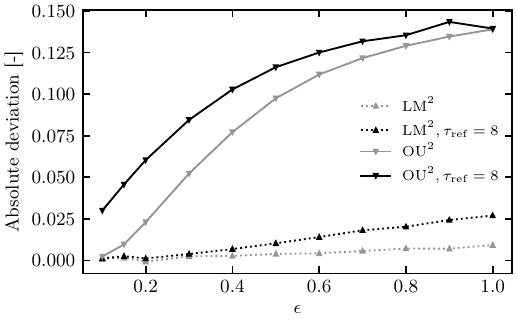}}
	\caption{Absolute deviation of the exact results of the Ising
		model for the absolute magnetization $m$ at the inverse
		critical temperature for the sign-dependent processes in
		dependence of $\epsilon$. The curves converge for all
		methods and all dynamics for smaller values of $\epsilon$ to
		the results of the Metropolis algorithm. The rate of
		convergence differs and signifies dependencies on the
		properties of the model, the intrinsic parameters and the
		update dynamics. Minor differences of the
		$\textnormal{OU}^2$ process occur due to finite time steps
		in the simulation.\label{fig:isiingepsdep}}
\end{figure}

The upper row of Figure~\ref{fig:refractory} shows the absolute value
of the magnetization for the Ising model with a vanishing external
magnetic field for the dynamics without and with a refractory
mechanism. The observables are computed for the $\textnormal{LM}^2$
and the $\textnormal{OU}^2$ process for different values of $\epsilon$
and for the $\textnormal{LM}^{1\textnormal{F}}$ and the
$\textnormal{OU}^{1\textnormal{F}}$ process. In
Figure~\ref{fig:errorall}, the deviation of the derived inverse critical
temperatures is plotted in dependency of $\epsilon$ for the processes
without a refractory mechanism. Figure~\ref{fig:isiingepsdep}
illustrates the convergence of the considered processes for vanishing
$\epsilon$. The resulting deviations reflect the magnitude of errors
in the representation of the activation function. This reinforces,
together with the results of Figure~\ref{fig:refractory}, the argument
that already small changes in the activation function can lead to
large deviations in the resulting observables. This argument also
explains the partially worse performance of the processes for the
dynamics with a refractory mechanism.

The observables exhibit for the $\textnormal{LM}^2$ and the
$\textnormal{OU}^2$ process the same tendency as the results for the
$4$-state clock model, despite their different sources of errors. The
equilibrium distributions are shifted to smaller values of $\beta$ as
described in the discussion of Section~\ref{sec:clockmodel}. As
before, the shift grows with larger values of $\beta$ and of
$\epsilon$. The similar behavior of the $\textnormal{OU}^2$ process
can be justified by the similar trend in the deviation of the
activation functions of the two processes. The higher rate of
convergence of the $\textnormal{LM}^2$ process is a result of the
different source of errors.

The comparison of the Kullback-Leibler divergence of the different
processes in the lower row of Figure~\ref{fig:refractory} reinforces the
better representation of the logistic distribution by the
sign-dependent processes and illustrates again the dependency on
$\epsilon$.

\section{Conclusions and Outlook}
\label{sec:concl}

In the present work we have introduced the discrete Langevin machine and discrete Langevin dynamics~\eq{eq:updaterulesimple}; see in particular Sections~\ref{sec:langevindynamicsdiscrete},~\ref{sec:langevinmachine} and~\ref{sec:wholepicture}.

The introduced dynamics paves the way for possible new
applications and the discovery of new physics. This includes, for
example, a formulation of Langevin dynamics for discrete systems with
respect to a possible computation of Hamiltonians with complex
contributions, similar to complex Langevin
dynamics~\cite{Aarts2008,Aarts2012}. A further interesting task is to
investigate the network architecture of the sign-dependent
Ornstein-Uhlenbeck process ($\textnormal{OU}^2$), given by the
dynamics~(\ref{eq:dynOU}) in Section~\ref{sec:neuromorphichardware},
with its particularity of a self-interacting term and resulting
fluctuating dynamics, on a neuromorphic hardware system.

The numerical analysis of different abstractions of a LIF network in
Section~\ref{sec:applications} demonstrates that the network
architecture of the sign-dependent discrete Langevin machine
($\textnormal{LM}^2$) and the $\textnormal{OU}^2$ process is suitable
for an exact computation of correlation functions of Boltzmann
distributed systems. This applies to both, a discrete two-state system
with uncorrelated noise and a continuous system with autocorrelated
noise. The numerical results show that an exact implementation of the
logistic distribution or at least a correct estimation of errors is
crucial to obtain quantitative exact observables.

It remains to be seen whether this statement is also sufficient and
valid for nontrivial PSP shapes, as a last step towards LIF
sampling. In particular, one has to analyze the impact of marginal
deviations to the activation function on observables of larger and
more complex systems than the one considered in this work. Moreover,
one may ask whether an exact representation of an activation function
with a self-interacting term is sufficient to also obtain reliable and
accurate results for interacting neurons independent of the
interaction kernel, i.e., the postsynaptic potential, respectively. In
other words: Is it possible to extend findings for a single
self-interacting neuron to a general complex interacting
system. These questions are postponed to future work. Either way, we
expect that the existence of a self-interacting contribution in the
$\textnormal{OU}^2$ process helps to better compensate arising
nonlinearities of the neuromorphic hardware.

In summary, the potentially more accurate implementation of Boltzmann
machines by the dynamics~(\ref{eq:langevinmachine}) and~(\ref{eq:dynOU}) represents a further step towards an integration of deep
learning and neuroscience~\cite{Marblestone, Emre, Leng}. We believe
that the present work offers a tool for a better comparison of classical
artificial networks and neuromorphic networks.

So far, the statistical properties of the introduced sign-dependent processes depend on an exact implementation of the underlying equations. Further, the sign-dependent Ornstein-Uhlenbeck process might suffer from large correlation times. These properties limit a broad application of the discovered processes on a wider class of models in biology and further stochastic dynamics at first sight. It is unclear which impact additional nonlinearities have on the characteristics of the processes. These nonlinearities include, for example, nontrivial PSP shapes or a different kind of noise. Therefore, it is subject to future work to study possibilities to embed the introduced dynamics into a wider class of stochastic processes.

Currently, the considered dynamics are restricted to additive Gaussian noise and are built upon sampling with leaky integrate-and-fire neurons as a neuron model. This applies also to the discussed mapping of a process with a discrete state space onto a system with continuous dynamics, as discussed in Section~\ref{sec:wholepicture}. An approach for sampling-based Bayesian spiking inference has been introduced recently in~\cite{Dold2019}. Their sampling approach works without any source of external noise but is driven instead by activities of neighboring sampling spiking neurons. It is interesting whether similar results can be observed for Bayesian inference based on the introduced sign-dependent dynamics in this work.

There are plenty of further interesting open questions. These concern, for example, different kinds of sources for noise or the inspection of a transferability on other underlying neuron models. Another important aspect is the integration and embedding of the introduced dynamics into similar existing dynamics, also with respect to other areas of applications. This allows a deeper understanding of interrelations and similarities of the developed dynamics with regard to systems of stochastic differential equations, in general. The modeling of population dynamics represents one of such areas of application. Multiplicative noise sources are often introduced in those models to imitate the stochastic impact of the environment. For example, the Verhulst model is considered in~\cite{Hamada1981} as a Langevin equation with a model-related drift term and multiplicative noise. In~\cite{Spagnolo2002, Spagnolo2004, Spagnolo2004A}, noisy systems of Lotka-Volterra equations are studied in a time-discrete description on a coupled map lattice as well as in their continuous form in time. Counterintuitive and interesting phenomena originate from the additional noise sources that range from the formation of spatiotemporal patterns of species to noise delayed spatial extinction~\cite{Spagnolo2002, Spagnolo2004, Spagnolo2004A} and noise-induced phase transitions~\cite{Hamada1981}.

The mathematical structure behind coupled map lattices is very similar to the evolution of the membrane potential of LIF neurons in discrete time with additional nontrivial interacting terms, which are based on the interactions of different neurons (see equations~(\ref{eq:ornsteinboltzmann}) and~(\ref{eq:dynOU})). Coupled map lattices~\cite{Kaneko1992} describe a dynamical system in discrete space and discrete time, but with a continuous state variable. In future work, we want to analyze whether there exist similar mappings as derived in this work (see equation~(\ref{eq:mapping})) for the dynamics of coupled map lattices. It might also be possible to transfer findings from these research areas onto the here considered dynamics and vice versa. A brief history of excitable map-based neurons and neural networks is given in~\cite{Girardi2013}, for example. Analogies might also be find in more related dynamics like the FitzHugh-Nagumo neuron model~\cite{FitzHugh1955, Nagumo1962} with an additional stochastic noise source. A representation of LIF neurons based on a stochastic Fitzhugo-Nagumo neural model is considered in~\cite{Yamakou2019}. In~\cite{Zaks2005}, collective dynamics of a noisey FitzHugh-Nagumo oscillator are studied. Critical phenomena and noise-induced phase transitions on classical random networks are provoked by shot noise in~\cite{Lee2014}. We expect that a detailed analysis of all these approaches together with the derived dynamics in this work will result in many interesting phenomena. This holds for simulations as well as for actual implementations on the neuromorhic hardware system.

\section*{Acknowledgements}

We thank M. A. Petrovici, A. Baumbach, and K. Meier for discussions and
collaboration on related subjects. This work is supported by the Deutsche Forschungsgemeinschaft (DFG, German Research Foundation) under Germany's Excellence Strategy EXC 2181/1 - 390900948 (the Heidelberg STRUCTURES Excellence Cluster).

\appendix
\section{Transition probability of the Langevin equation}
\label{sec:transitionprobability}

The transition probabilities of the discrete Langevin equation are
computed, in the following. These are used in Section~\ref{sec:equivalence} for an
interpretation of the dynamics as a standard Monte Carlo
algorithm. Starting from the discrete Langevin equation
\begin{equation}
\label{eq:discreteLangevin}
\phi'=\phi-\epsilon\frac{\delta S}{\delta \phi_x}+\sqrt{\epsilon}\eta\,,
\end{equation}
with: $\phi:=\phi(\tau)$ and $\phi':=\phi(\tau+\epsilon)$, it is
straightforward to compute the transition probabilities of an
infinitesimal change,
\begin{equation}
W(\phi\to\phi')=\frac{1}{\sqrt{2\epsilon}}\varphi\left(\frac{\phi'-\phi}{\sqrt{2\epsilon}}
+\sqrt{\frac{\epsilon}{2}}\frac{\delta S}{\delta \phi}\right)\,.
\end{equation}
Inserting the standard normal distribution
$\varphi(x)=\frac{1}{\sqrt{2\pi}}\exp\left[-\frac{1}{2x^2}\right]$ and
computing the square in the exponent one obtains
\begin{align}
W(\phi\to\phi')&=\frac{1}{\sqrt{4\pi\epsilon}}\exp\left[-\frac{1}{2}\left(\frac{\phi'
	-\phi}{\sqrt{2\epsilon}}+\sqrt{\frac{\epsilon}{2}}\frac{\delta S}{\delta \phi}\right)^2\right]=\nonumber\\[1ex]
&=\varphi\left(\frac{\phi'-\phi}{\sqrt{2\epsilon}}\right)\exp\left[-\frac{\phi'
	-\phi}{2}\frac{\delta S}{\delta \phi}+\mathcal{O}(\epsilon)\right]\,.
\end{align}
With the identifications $\delta \phi\simeq\phi'-\phi$ and
$\delta S \simeq S(\phi') - S(\phi)$, this can be further simplified
to
\begin{align}
W(\phi\to\phi')&=\nonumber\\[1ex]
=\frac{1}{\sqrt{2\epsilon}}&\varphi\left(\frac{\phi'-\phi}{\sqrt{2
		\epsilon}}\right)\exp\left[-\frac{S(\phi') - S(\phi)}{2} + \mathcal{O}(\epsilon)\right]\,.
\end{align}
Apparently, the transition probability satisfies the detailed balance
equation since the first factor is symmetric to an exchange of $\phi'$
and $\phi$,
\begin{equation}
\frac{W(\phi\to\phi')}{W(\phi'\to\phi)}=\exp\left[-(S(\phi')- S(\phi))\right]\,.
\end{equation}
\begin{figure*}[htbp]
	\centering\resizebox{0.318\textwidth}{!}{\includestandalone[mode=buildnew]{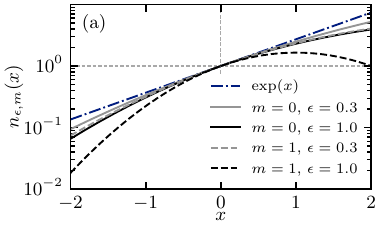}}\hfill\resizebox{0.324\textwidth}{!}{\includestandalone[mode=buildnew]{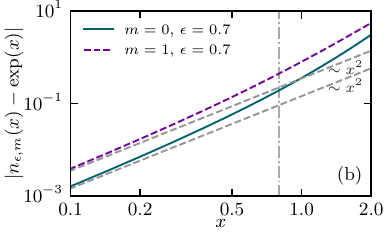}}\hfill\resizebox{0.332\textwidth}{!}{\includestandalone[mode=buildnew]{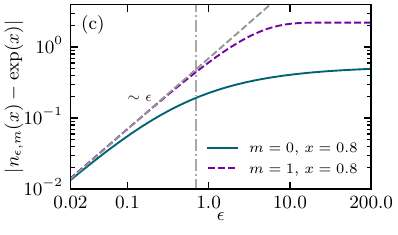}}
	\caption{Illustrations regarding the limit $\lim\limits_{\epsilon\to 0} n_{\epsilon, m}(x)$ of relations~(\ref{eq:cumexp}) and~(\ref{eq:cumexpm}) for $m=0$ and $m=1$: (a)~Comparison to the exponential function. (b)~Dependency on $x$ for a fixed $\epsilon=0.7$. (c)~Dependency on $\epsilon$ for a fixed value of $x=0.8$. (a)-(c): The vertical lines in~(b) and~(c) indicate the respective fixed value of $\epsilon$ and $x$. In general, the limit with the cumulative Gaussian distribution ($m=0$) has a lower deviation for equal values of $\epsilon$ then the limit with the Gaussian distribution ($m=1$).}
	\label{fig:convergenceexp}
\end{figure*}

\section{Relation between the cumulative normal distribution and the exponential function}

Turning to discrete states, the normal distribution as a density
probability distribution transforms to differences of cumulative
normal distributions.

To be able to define an update formalism with a Gaussian noise term,
we need a relation between the exponential function of the transition
probability and the cumulative normal distribution. Such a relation
exists and is given by
\begin{equation}
\label{eq:cumexp}
\lim\limits_{\epsilon\to 0} n_{\epsilon, 0}(x)=
\lim\limits_{\epsilon\to 0}\frac{\Phi\left(-\frac{1}{\sqrt{\epsilon}}+\sqrt{\epsilon}\frac{x} {\lambda_\epsilon}\right)}{\Phi(-\frac1{\sqrt{\epsilon}})}=\exp(x)+\mathcal{O}(\epsilon x^2)\,,
\end{equation}
with a scaling factor
\begin{equation}
\label{eq:lambda}
\lambda_\epsilon = \frac{\sqrt{\epsilon}\varphi\left(-\frac1{\sqrt{\epsilon}}\right)}{\Phi\left(-\frac1{\sqrt{\epsilon}}\right)}\,,
\end{equation}
and the cumulative normal distribution $\Phi(x)=\int_{-\infty}^{x}dt \frac{1}{\sqrt{2\pi}}\exp\left[-\frac{1}{2x^2}\right]$. As shown in Appendix~\ref{sec:relationcumnormalandexp}, the denominator in the limit corresponds to a scaling of the zero order term of the Taylor expansion of the cumulative normal distribution. The rescaling of $x$ in the argument corrects the first order term. The resulting order of accuracy is therefore of second order in $\sqrt{\epsilon}$.

The relation can be extended to the $m$-th derivative of the cumulative distribution with $m>0$, according to
\begin{align}
\label{eq:cumexpm}
\lim\limits_{\epsilon\to 0}&n_{\epsilon,m}(x)=\lim\limits_{\epsilon\to 0}\frac{\frac{\partial^{m}}{\partial t^{m}}\Phi\left(-\frac{1}{\sqrt{\epsilon}}+\sqrt{\epsilon} t\right)\bigg|_{t=x/\sigma_{m, \epsilon}}}{\frac{\partial^{m}}{\partial t^{m}}\Phi(-\frac1{\sqrt{\epsilon}}+\sqrt{\epsilon} t)\big|_{t=0}}=\nonumber\\[1ex]
&\quad\quad\quad\quad\quad\quad\quad\quad=\exp(x)+\mathcal{O}(\epsilon x^2)\,,
\end{align}
where the scaling factor $\sigma_{m,\epsilon}$ is defined as
\begin{equation}
\sigma_{m,\epsilon}=-\frac{\sqrt{\epsilon}\textnormal{He}_m\left(-\frac{1}{\sqrt{\epsilon}}\right)}{\textnormal{He}_{m-1}\left(-\frac{1}{\sqrt{\epsilon}}\right)}\,,
\end{equation}
and where $\textnormal{He}_m(x)$ denote the $m$-th probabilists' Hermite polynomials. Figure~\ref{fig:convergenceexp} illustrates the dependence on $\epsilon$ and on $x$ of the two relations~(\ref{eq:cumexp}) and~(\ref{eq:cumexpm}).

For $m=1$, this corresponds to a similar computation as in Appendix~\ref{sec:transitionprobability} for the Langevin equation of continuous systems. The scaling factor of $x$ becomes $1$ and the resulting identity simplifies to
\begin{equation}
\label{eq:limitphi}
\lim\limits_{\epsilon\to 0}n_{\epsilon, 1}(x)=\lim\limits_{\epsilon\to 0}\frac{\varphi\left(-\frac{1}{\sqrt{\epsilon}}+\sqrt{\epsilon}x\right)}{\varphi\left(-\frac{1}{\sqrt{\epsilon}}\right)}=\exp(x)+\mathcal{O}(\epsilon x^2)\,.
\end{equation}

\section{Derivation of the relation between the cumulative normal distribution and the exponential function}
\label{sec:relationcumnormalandexp}

The relations~(\ref{eq:cumexp}) and~(\ref{eq:cumexpm}) are derived. For reasons of readability, $\sqrt{\epsilon}$ is abbreviated by $\varepsilon$ and the shorthand notation $\frac{\partial^n}{\partial x^n}=\partial^n$ is used in the following.

We start with a Taylor series around $x = 0$ of the $m$-th derivative of the cumulative Gaussian contribution,
\begin{equation}
\label{eq:taylor}
\partial^m \Phi\left(-\frac{1}{\varepsilon}+\varepsilon x\right)=\sum_{n=0}^{\infty}\frac1{n!}\partial^{n+m}\Phi\left(-\frac{1}{\varepsilon}+\varepsilon x\right)\bigg|_{x = 0} x^n\,.
\end{equation}
The following important identity between the cumulative Gaussian
distribution $\Phi(x)$ and the probabilists' Hermite polynomials
$\textnormal{He}_n(x)$ is useful for an evaluation of the Taylor
expansion:
\begin{align}
\label{eq:derivativePhi}
  \partial^{n+m}\Phi\left(-\frac{1}{\varepsilon}+\varepsilon x\right)=\quad\quad&\nonumber\\[1ex]
  =(-\varepsilon)^{n+m-1}\textnormal{He}_{n+m-1}&\left(-\frac{1}{\varepsilon}+
                                                  \varepsilon x\right)\partial\Phi\left(-\frac{1}{\varepsilon}+\varepsilon x\right)\,,
\end{align}
for $n>0$. Since this relation holds only for $n>0$, the cases for
$m=0$ and $m>0$ have to be treated separately, which coincides with
the relations~(\ref{eq:cumexp}) and~(\ref{eq:cumexpm}).

\subsection{Evaluation for $m=0$}

Inserting relation~(\ref{eq:derivativePhi}) into the Taylor
expansion~(\ref{eq:taylor}) and setting $m=0$, one yields
\begin{align}
  &\Phi\left(-\frac{1}{\varepsilon}+\varepsilon x\right)=\Phi\left(
    -\frac1\varepsilon\right)+\sum_{n=1}^{\infty}\frac1{n!}(-\varepsilon)^{n-1}\times\nonumber\\[1ex]
  &\quad\times\textnormal{He}_{n-1}\left(-\frac{1}{\varepsilon}+
    \varepsilon x\right)\bigg|_{x = 0}\partial\Phi\left(-\frac{1}{\varepsilon}+\varepsilon x\right)\bigg|_{x = 0} x^n\nonumber=\\[1ex]
  &\quad=\Phi\left(-\frac1\varepsilon\right)+\nonumber\\[1ex]
  &\quad+\sum_{n=1}^{\infty}\frac1{n!}(-1)^{n-1}\varepsilon^n\textnormal{He}_{n-1}
    \left(-\frac1\varepsilon\right)\varphi\left(-\frac1\varepsilon\right) x^n\,.
\end{align}
A comparison with the Taylor expansion of
$\exp(x)=1+x+\mathcal{O}(x^2)$ shows that the first two terms in the
above expression can be fixed by a division of the entire equation by
$\Phi\left(-\frac1\varepsilon\right)$ and an additional rescaling of
$x$ by
\begin{equation}
  \lambda(\varepsilon)=\frac{\varepsilon \varphi\left(-\frac1
      \varepsilon\right)}{\Phi\left(-\frac1\varepsilon\right)}\,.
\end{equation}
This gives
\begin{align}
  \frac{\Phi\left(-\frac{1}{\varepsilon}+\varepsilon \frac{x}{
  \lambda(\epsilon)}\right)}{\Phi\left(-\frac1\varepsilon\right)}=&\nonumber\\[1ex]
  =1+x+\sum_{n=2}^{\infty}\frac1{n!}&\frac{(-1)^{n-1}
                                      \varepsilon^{n-1}\textnormal{He}_{n-1}
                                      \left(-\frac1\varepsilon\right)}{\lambda(\varepsilon)^{n-1}} x^n\,.
\end{align}
It remains to show that the fractional factor converges to $1$ for
$\epsilon\to0$ and for arbitrary values of $n$. This is done in two
steps. First, we argue that
$\lim\limits_{\varepsilon\to0}\lambda(\varepsilon) = 1 +
\mathcal{O}(\varepsilon^2)$ and, second, a limit is derived for the
fractional factor.

The limit of $\lim\limits_{\varepsilon\to0}\lambda(\varepsilon)$ can
be derived by showing the identity that
$\lim\limits_{\varepsilon\to0}\Phi\left(\frac{1}{\varepsilon}\right) =
\lim\limits_{\varepsilon\to0}\varphi\left(\frac{1}{\varepsilon}\right)$. By
the substitution $u:=\frac{1}{x}$ and a subsequent partial
integration, one finds that a second order term in $x = \frac{1}{u}$
vanishes and arrives directly at the identity, which entails that
\begin{equation}
\label{eq:limitlambda}
\lim\limits_{\varepsilon\to0}\lambda(\varepsilon) = 1 + \mathcal{O}(\varepsilon^2)\,.
\end{equation}
Since the highest order term of the $n$-th probabilists' Hermite polynomial equals $x^n$, it can be directly concluded that
\begin{equation}
\label{eq:limitHermite}
\lim\limits_{\varepsilon\to 0}\varepsilon^n\textnormal{He}_n\left(-\frac1\varepsilon\right)=(-1)^n+\mathcal{O}(\varepsilon^n)\,.
\end{equation}
Using the Taylor expansion
\begin{equation}
\frac{1}{\left(1+x\right)^{n-1}} = 1 - (n-1)x + \mathcal{O}(x^2)\,,
\end{equation}
and inserting the two limits~(\ref{eq:limitlambda}) and~(\ref{eq:limitHermite}), one arrives at the following limit for the fractional factor:
\begin{equation}
\lim\limits_{\varepsilon\to 0}\frac{(-1)^{n-1}\varepsilon^{n-1}\textnormal{He}_{n-1}\left(-\frac1\varepsilon\right)}{\lambda(\varepsilon)^{n-1}}=1+\mathcal{O}(\varepsilon^2)\,.
\end{equation}
The final limit between the cumulative normal distribution and the exponential function can be stated with the corresponding order of accuracy,
\begin{equation}
n_{\varepsilon^2, 0}(x)=\frac{\Phi\left(-\frac{1}{\varepsilon}+\varepsilon \frac{x}{\lambda(\varepsilon)}\right)}{\Phi\left(-\frac1\varepsilon\right)}=\exp(x)+\mathcal{O}(\varepsilon^2x^2)\,.
\end{equation}
The existence of this limit can also be proven by applying L'Hôpital's rule to relation~(\ref{eq:limitphi}) as shown in~\cite{Yi2010}.

\subsection{Evaluation for $m>0$}

Proceeding similarly as for the case of $m=0$, The Taylor expansion can be written as
\begin{align}
\partial^m \Phi&\left(-\frac{1}{\varepsilon}+\varepsilon x\right)=\partial^m\Phi\left(-\frac1\varepsilon+\varepsilon x\right)\bigg|_{x=0}+\sum_{n=1}^{\infty}\frac1{n!}\times\nonumber\\[1ex]
&\times(-1)^{n+m-1}\varepsilon^{n+m}\textnormal{He}_{n+m-1}\left(-\frac1\varepsilon\right)\varphi\left(-\frac1\varepsilon\right) x^n\,.
\end{align}
The fixing of the first two order terms of the exponential function leads to
\begin{align}
&\frac{\partial^m\Phi\left(-\frac{1}{\varepsilon}+\varepsilon x\right)\big|_{x=x/\sigma_m(\epsilon)}}{\partial^m\Phi\left(-\frac1\varepsilon+\varepsilon x\right)\big|_{x=0}}=1+x+\nonumber\\[1ex]
&+\sum_{n=2}^{\infty}\frac1{n!}\frac{(-1)^{n+m-1}\varepsilon^{n+m}\textnormal{He}_{n+m-1}\left(-\frac1\varepsilon\right)\varphi\left(-\frac1\varepsilon\right)}{(-1)^{m-1}\varepsilon^m\textnormal{He}_{m-1}\left(-\frac1\varepsilon\right)\varphi\left(-\frac1\varepsilon\right)\sigma_m(\varepsilon)^{n}} x^n=\nonumber\\[1ex]
&=1+x+\sum_{n=2}^{\infty}\frac1{n!}\frac{(-1)^n\varepsilon^{n}\textnormal{He}_{n+m-1}\left(-\frac1\varepsilon\right)}{\textnormal{He}_{m-1}\left(-\frac1\varepsilon\right)\sigma_m(\varepsilon)^{n}} x^n\,,
\end{align}
where we have evaluated the expression $\partial^m\Phi\left(-\frac1\varepsilon+\varepsilon x\right)\bigg|_{x=0}$ on the right-hand side and where the scaling factor $\sigma_m(\varepsilon)$ is given by
\begin{equation}
\sigma_m(\varepsilon)=-\frac{\varepsilon\textnormal{He}_m\left(-\frac{1}{\varepsilon}\right)}{\textnormal{He}_{m-1}\left(-\frac{1}{\varepsilon}\right)}\,.
\end{equation}
Again, the asymptotic behavior of the fractional factor has to be computed for $\epsilon\to0$. From the highest order term of the probabilists' Hermite polynomial, it can be deduced that
\begin{equation}
\lim\limits_{\varepsilon\to 0}\frac{\varepsilon^n\textnormal{He}_{n+m-1}\left(-\frac1\varepsilon\right)}{\textnormal{He}_{m-1}\left(-\frac1\varepsilon\right)}=(-1)^n + \mathcal{O}(\epsilon^n)\,.
\end{equation}
For $n=1$, this corresponds to $(-1)$ times the scaling factor $\sigma(\varepsilon)$, therefore,
\begin{equation}
\lim\limits_{\varepsilon\to 0}\sigma(\varepsilon)=1+\mathcal{O}(\varepsilon^n)\,.
\end{equation}
It can be derived with the same arguments as for the case of $m=0$, that the fractional factor converges to $1$ and that the limit with its order of accuracy is given by
\begin{equation}
n_{\varepsilon^2, m}=\frac{\partial^m\Phi\left(-\frac{1}{\varepsilon}+\varepsilon x\right)\big|_{x=x/\sigma_m(\epsilon)}}{\partial^m\Phi\left(-\frac1\varepsilon+\varepsilon x\right)\big|_{x=0}}=\exp(x)+\mathcal{O}(\varepsilon^2x^2)\,.
\end{equation}
\section{Statistical properties of the sign-dependent Ornstein-Uhlenbeck process}
\label{sec:statsigndependent}

The transition probability of the sign-dependent Ornstein-Uhlenbeck process can be derived in the same manner as for Langevin dynamics in Appendix~\ref{sec:transitionprobability}. We start by considering the process~(\ref{eq:dynOU}) with discrete time steps,
\begin{align}
\label{eq:dynOUdiscrete}
u&_i' =u_i\nonumber\\[1ex] &+
\varepsilon\left[\frac{\textnormal{sign}(u_i)}{\sqrt{\epsilon}}+\frac{\sqrt{\epsilon}}{2}\left(\sum_{\textnormal{syn} j} W_{ij}z_j
+ b_i\right)-u_i\right]\nonumber\\[1ex]
&+\sqrt{2\varepsilon}\tilde{\eta}\,,
\end{align}
with $\theta=1$ and $\sigma=\sqrt{2}$. The $t$-dependency is hidden for simplicity by introducing $u_i:=u_{i, \textnormal{eff}}(t)$ and $u_i':=u_{i, \textnormal{eff}}(t + \varepsilon)$ as well as $z_j:=z_j(t)$. We define the term in the squared brackets as
\begin{equation}
\frac{\partial K(u_i)}{\partial u_i} := \frac{\textnormal{sign}(u_i)}{\sqrt{\epsilon}}+\frac{\sqrt{\epsilon}}{2}\left(\sum_{\textnormal{syn} j} W_{ij}z_j
+ b_i\right)-u_i\,.
\end{equation}
This definition allows a one-to-one comparison between the discrete update equation~(\ref{eq:dynOUdiscrete}) and the discrete Langevine equation~(\ref{eq:discreteLangevin}). The transition probability for the sign-dependent Ornstein-Uhlenbeck process can be expressed based on this comparison by
\begin{align}
W&(u_i\to u_i')=\nonumber\\[1ex]
&=\frac{1}{\sqrt{2\varepsilon}}\varphi\left(\frac{u_i'-u_i}{\sqrt{2\varepsilon}}\right)\exp\left[\frac{K(u_i') - K(u_i)}{2} + \mathcal{O}(\varepsilon)\right]\,,
\end{align}
with
\begin{equation}
K(u_i)=\frac{|u_i|}{\sqrt{\epsilon}}+\frac{\sqrt{\epsilon}}{2}\left(\sum_{\textnormal{syn} j} W_{ij}z_j
+ b_i\right)u_i-\frac{u_i^2}{2}\,.
\end{equation}
The equilibrium distribution can be inferred from the detailed balance equation and is given by
\begin{equation}
P_\textnormal{eq}(u_{i,\textnormal{eff}}) \propto \exp\left[K(u_i)\right]\,.
\end{equation}
This result enables a derivation of the probability distributions of $P_{\textnormal{OU}^2}(z_i=1)$ and $P_{\textnormal{OU}^2}(z_i=-1)$, where we use that $z_i=\Theta\left[u_{i,\textnormal{eff}}\right]$. One obtains for the distributions
\begin{align}
P_{\textnormal{OU}^2}&(z_i=1)=\int_{0}^{\infty}
P_\textnormal{eq}(u_{i,\textnormal{eff}})\textnormal{d}u_{i,\textnormal{eff}}\propto\nonumber\\[1ex]&\propto\exp\left[\frac{\left(\epsilon m_i - 2\right)^2}{8\epsilon}\right]\Phi\left[\frac{1}{\sqrt{\epsilon}} - \frac{\sqrt{\epsilon}}{2}m_i\right]
\end{align}
and
\begin{align}
P_{\textnormal{OU}^2}&(z_i=-1)=\int_{-\infty}^{0}
P_\textnormal{eq}(u_{i,\textnormal{eff}})\textnormal{d}u_{i,\textnormal{eff}}\propto\nonumber\\[1ex]&\propto\exp\left[\frac{\left(\epsilon m_i + 2\right)^2}{8\epsilon}\right]\Phi\left[\frac{1}{\sqrt{\epsilon}} + \frac{\sqrt{\epsilon}}{2}m_i\right]\,,
\end{align}
where we defined the total input of a neuron $i$ as
\begin{equation}
m_i:=-\sum_{\textnormal{syn} j} W_{ij}z_j
- b_i\,.
\end{equation}
Since we consider only two states for $z_i$, $P_{\textnormal{OU}^2}(z_i=1)$ can be reformulated based on the previous definition. This leads to
\begin{equation}
P_{\textnormal{OU}^2}(z_i=1)=\frac{1}{1+\exp\left[\alpha_{\epsilon}(m_i)\times m_i\right]}\,.
\end{equation}
The correction factor $\alpha(m_i)$ is given by
\begin{equation}
\alpha(m_i):=1 + \frac{1}{m_i}\log\left[\frac{ \Phi\left(\frac{1}{\sqrt{\epsilon}} + \frac{\sqrt{\epsilon}}{2}m_i\right)}{
\Phi\left(\frac{1}{\sqrt{\epsilon}} - \frac{\sqrt{\epsilon}}{2}m_i\right)}\right]\,.
\end{equation}
The correction factor converges in the limit of $\epsilon\to0$ to one,
\begin{equation}
\lim\limits_{\epsilon\to 0}\alpha_{\epsilon}(m_i) = 1\,.
\end{equation}
\section{Derivation of the dynamics of the Langevin machine}
\label{sec:derivationlangevinmachine}

We start from the energy of the Boltzmann machine
\begin{equation}
E = -\sum_{i<j} W_{ij} z_i z_j - \sum_i b_i z_i\,,
\end{equation}
with a total input $m_i:=-\sum_{j} W_{ij} z_j - b_i$ for a neuron $i$. The possible resulting energy differences for a change of the state of the neuron are given by
\begin{align}
\Delta E(z_i'=1,z_i=0) &= m_i\,,\nonumber\\[1ex]
\Delta E(z_i'=0,z_i=1) &= -m_i\,.
\end{align}
The proposed state for a two-state system always corresponds to the other state. This results in the following two update rules for a transition from $z_i=0\to1$ and $z_i=1\to0$:
\begin{align}
z_i'= \Theta\left[-1-\frac{\epsilon}{2\lambda_\epsilon}m_i+\sqrt{\epsilon}\tilde{\eta_i}^T\right]\,,\quad\textnormal{for}\,\quad z_i=0\to 1\,,\nonumber\\[1ex]
z_i'= \Theta\left[1-\frac{\epsilon}{2\lambda_\epsilon}m_i+\sqrt{\epsilon}\tilde{\eta_i}^T\right]\,,\quad\textnormal{for}\,\quad z_i=1\to 0\,.
\end{align}
Taking the current state into account, the relations can be merged into a common update rule,
\begin{equation}
z_i' = \Theta\left[\left(2z_i-1\right)-\frac{\epsilon}{2\lambda_\epsilon}m_i+\sqrt{\epsilon}\tilde{\eta_i}^T\right]\,.
\end{equation}
After a division by $\sqrt{\epsilon}$ and a rearranging of the summands, one arrives at the following update rule for the Langevin machine,
\begin{equation}
z'_i=\Theta\left[\frac{2}{\sqrt{\epsilon}}z_i+\sum_{j}\frac{\sqrt{\epsilon}}{2\lambda_\epsilon}W_{ij}z_j+\frac{\sqrt{\epsilon}}{2\lambda_{\epsilon}}b_i-\frac1{\sqrt{\epsilon}}+\tilde{\eta_i}^T\right]\,.
\end{equation}
By the identifications
\begin{equation}
W'_{ii} = \frac{2}{\sqrt{\epsilon}}\,,\quad
W'_{ij} = \frac{\sqrt{\epsilon}}{2\lambda_\epsilon}W_{ij}\,,\quad
b'_i = \left(\frac{\sqrt{\epsilon}}{2\lambda_{\epsilon}}b_i-\frac1{\sqrt{\epsilon}}\right)\,,
\end{equation}%
the update rule can be written as
\begin{equation}
z'_i=\Theta\left[W'_{ii}z_i+\sum_{j} W'_{ij}z_j + b'_i + \tilde{\eta}^T\right]\,.
\end{equation}

\section{Interactions on the neuromorphic hardware system}
\label{sec:interaction}

The interaction between neurons is nontrivial. The postsynaptic
potential (PSP) corresponds to the input potential of a connected
neuron in case of firing. As shown in~\cite{Petrovici2016}, the
interaction term can be approximated by
\begin{equation}
\mu_{i}(t)^\textnormal{interaction}=\sum_{\textnormal{syn}\, j}\sum_{\textnormal{spk}\, s} A_{ij} \kappa(t, t_{s,j})\,,
\end{equation}
with $A_{ij}=\frac{w_{ij} \left(E_{ij}^\textnormal{rev}-\langle
    u_\textnormal{eff}\rangle \right)}{\langle
  g_\textnormal{tot}\rangle}$ and where $t_{s,j}$ is the time of the
last spike. $\kappa(t, t_{s,j})$ describes the PSP shape and depends
in general on the time constants: $\tau_\textnormal{ref}$,
$\tau_\textnormal{syn}$ and $\tau_\textnormal{eff}$.

The actual PSP shape has the following form~\cite{Petrovici2016}: 
\begin{equation}
\kappa(t, t_{s,j}) = \frac{\exp\left[-\frac{t-t_{s,i}}{\tau_\textnormal{eff}}\right]-\exp\left[-\frac{t-t_{s,j}}{\tau_\textnormal{syn}}\right]}{\tau_\textnormal{eff}-\tau_\textnormal{syn}}\,.
\end{equation}
Weights $W_{ij}$ can be translated onto the neuromorphic system by the
assumption that the area under a PSP shape is equal to
$W_{ij}\tau_\textnormal{ref}\alpha$, where $\alpha$ is some scaling
factor,
\begin{equation}
W_{ij}\tau_\textnormal{ref}\alpha = \int_0^\textnormal{ref}A_{ij}\kappa(t, t_{s,j})\, \textnormal{d} t\,.
\end{equation}
Ideally, the PSP shape would have a rectangular form, 
\begin{equation}
\kappa(t, t_{s,j})^\textnormal{rect} = \Theta\left[t-t_{s,j}\right]-\Theta\left[t-t_{s,j}-\tau_\textnormal{ref}\right]\,.
\end{equation}
For $\tau_\textnormal{ref}\to 0$, the neuron $j$ is in the firing mode
as long as $u_j(t)>\vartheta_j$ and the interaction term turns to 
\begin{equation}
\mu_i(t)^\textnormal{interaction} = \sum_{\textnormal{syn} j} A_{ij} \Theta\left[u_j(t)-\vartheta_j\right]\,.
\end{equation}

\bibliography{literature}

\end{document}